\documentclass[twoside,11pt]{article}

\usepackage{blindtext}
\usepackage{jmlr2e}
\usepackage{microtype}
\usepackage{graphicx}
\usepackage{subfigure}
\usepackage{booktabs} 
\usepackage{amssymb}
\usepackage{enumitem}
\usepackage{hyperref}
\usepackage{amsmath}
\usepackage{color}
\usepackage{algorithm}
\usepackage{algorithmic}
\newcommand{\homo} {{\mathrm{H}}}
\newcommand{\myparagraph}[1]{\smallskip\noindent\textbf{#1}}

\usepackage{lastpage}
\jmlrheading{26}{2025}{1-\pageref{LastPage}}{10/23; Revised 12/24}{1/25}{23-1424}{Zuoyu Yan, Qi Zhao, Ze Ye, Tengfei Ma, Liangcai Gao, Zhi Tang, Yusu Wang and Chao Chen}
\ShortHeadings{Enhancing Graph Representation Learning with Localized Topological Features}{Yan, Zhao, Ye, Ma, Gao, Tang, Wang and Chen}
\editor{Sayan Mukherjee}

\firstpageno{1}

\begin{document}
%
\title{Enhancing Graph Representation Learning with Localized Topological Features}

\author{\name Zuoyu Yan$^1, ^5$ \email zuy4001@med.cornell.edu 
\AND \name Qi Zhao$^2$ \email qiz032@ucsd.edu 
\AND \name Ze Ye$^3$ \email ze.ye@stonybrook.edu 
\AND \name Tengfei Ma$^{3,*}$ \email Tengfei.Ma@stonybrook.edu  
\AND \name Liangcai Gao$^{1,*}$ \email glc@pku.edu.cn 
\AND \name Zhi Tang$^1$ \email tangzhi@pku.edu.cn 
\AND \name Yusu Wang$^{4,*}$ \email yusuwang@ucsd.edu 
\AND \name Chao Chen$^{3,*}$ \email chao.chen.1@stonybrook.edu \\
\addr $^1$ Wangxuan Institute of Computer Technology, Peking University \\
$^2$ Computer Science and Engineering Department, University of California, San Diego \\
$^3$ Department of Biomedical Informatics, Stony Brook University \\
$^4$ Hal{\i}c{\i}o\u{g}lu Data Science Institute, University of California, San Diego\\
$^5$ Weill Cornell Medicine, Cornell University\\
$^*$ Correspondence Author\\
}

\maketitle

\begin{abstract}
Representation learning on graphs is a fundamental problem that can be crucial in various tasks. 
Graph neural networks, the dominant approach for graph representation learning, are limited in their representation power. 
Therefore, it can be beneficial to explicitly extract and incorporate high-order topological and geometric information into these models. In this paper, we propose a principled approach to extract the rich connectivity information of graphs based on the theory of persistent homology. Our method utilizes the topological features to enhance the representation learning of graph neural networks and achieve state-of-the-art performance on various node classification and link prediction benchmarks. We also explore the option of end-to-end learning of the topological features, i.e., treating topological computation as a differentiable operator during learning. Our theoretical analysis and empirical study provide insights and potential guidelines for employing topological features in graph learning tasks.\\
\textbf{Keywords:} Persistent Homology, Topological Data Analysis, Graph Neural Network, Graph Representation Learning, Graph Isomorphism
\end{abstract}

\section{Introduction}
\label{sec:introduction}

Recent years have witnessed rapid development in graph representation learning.
Graph neural networks (GNNs) have become very popular for graph-structured data~\citep{zhang2020deep, wu2020comprehensive, ma2021deep}, and have achieved state-of-the-art performance. Existing GNNs learn task-relevant node embeddings through message passing and encoding \citep{kipf2016semi,velivckovic2018graph}. They utilize both nodes' intrinsic attributes and graph structure (e.g., node degree) for better representation learning~\citep{nickel2014reducing, zhao2017leveraging}.

\begin{figure}[bth!]
	\centering
	\vspace{-.1in}
	\includegraphics[width=0.6\columnwidth]{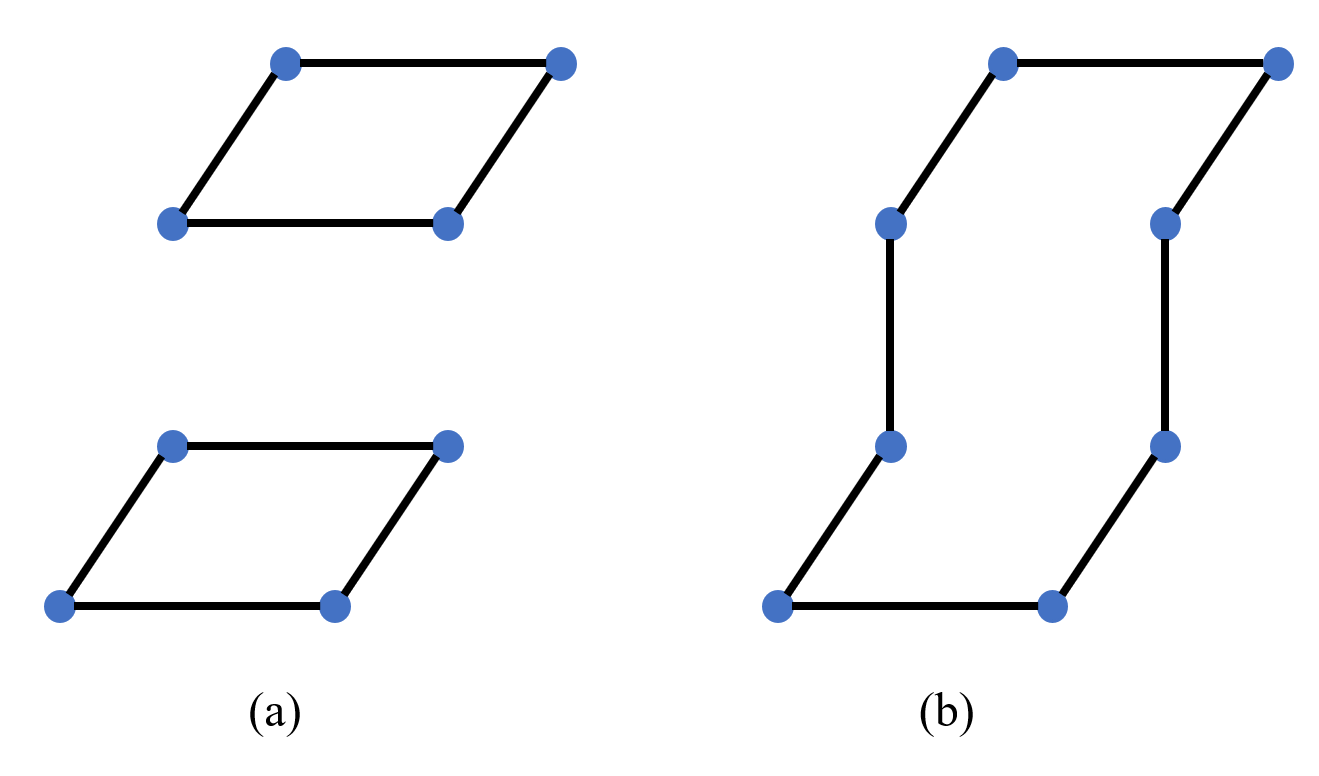}
	\vspace{-.2in}
	\caption{Two example graphs that 1-WL cannot distinguish. Topological information such as the number of cycles can differentiate them.}
	\label{fig:motiv}
	\vspace{-.1in}
\end{figure}

However, the expressive power of classic GNNs is limited; message-passing GNNs are at most as powerful as the Weisfeiler-Lehman test (the 1-st order WL-test, namely 1-WL)~\citep{weisfeiler1968reduction} in terms of distinguishing non-isomorphic graphs~\citep{xu2018powerful}. For instance, the two graphs in Figure~\ref{fig:motiv} cannot be distinguished by 1-WL. Thus they cannot be distinguished by classic message-passing GNNs either.
Even with structural information such as the node degree~\citep{kipf2016semi}, these two graphs still cannot be differentiated, since all nodes have the same degree.

\begin{figure*}[btp]
	\centering
	\subfigure[]{
		\begin{minipage}[t]{0.20\linewidth}
			\centering
			\includegraphics[width=\columnwidth]{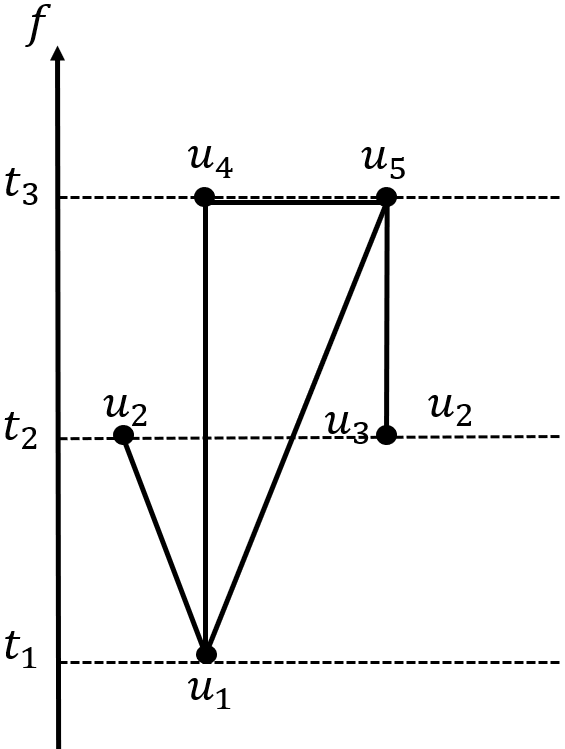}
		\end{minipage}%
	}%
	\subfigure[]{
		\begin{minipage}[t]{0.57\linewidth}
			\centering
			\includegraphics[width=\columnwidth]{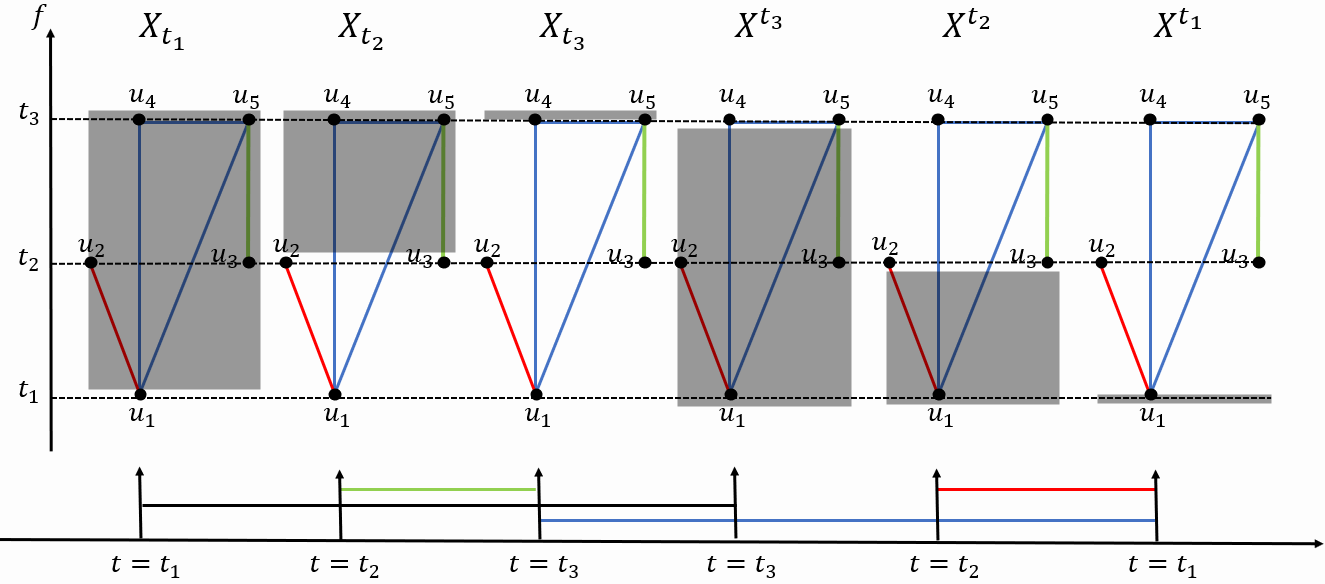}
		\end{minipage}%
	}%
	\subfigure[]{
		\begin{minipage}[t]{0.23\linewidth}
			\centering
			\includegraphics[width=\columnwidth]{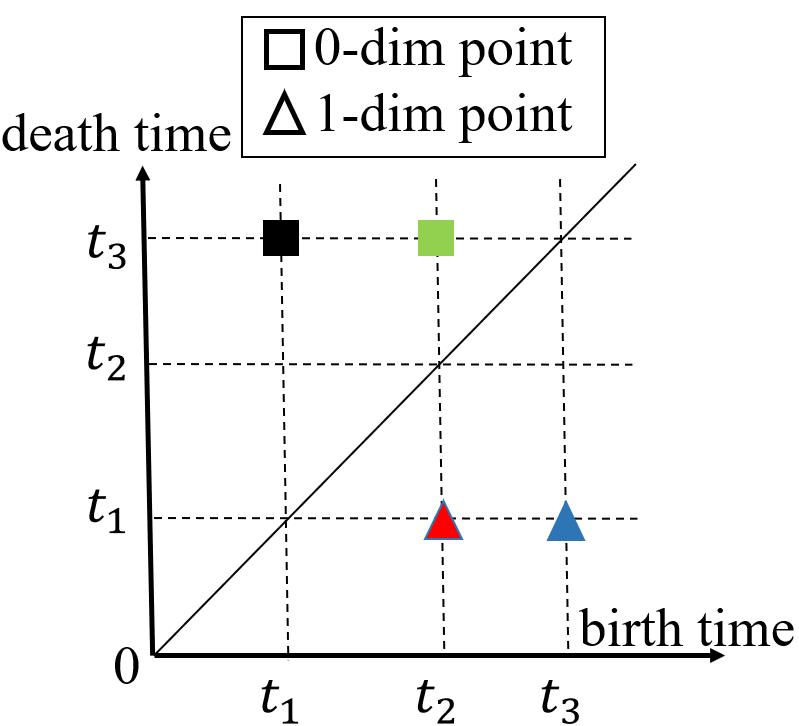}
		\end{minipage}
	}%
	\centering
	\vspace{-.1in}
	\caption{An illustration of extended persistent homology. (a) The input graph. It consists of 5 nodes, with filter values $f(u_1)=t_1$, $f(u_2) = t_2$, $f(u_3) = t_2$, $f(u_4) = t_3$, and $f(u_5) = t_3$. (b) The ascending and descending filtration of the input graph. The green, red, black, and blue bar correspond to the life span of the green part, the red part, the whole graph, and the blue loop in the graph respectively. The first three figures denote the ascending filtration, while the last three figures denote the descending filtration. In the ascending filtration, the filter function for edges is defined as $f(uv) = max(f(u),f(v))$, while in the descending filtration, $f(uv) = min(f(u),f(v))$. (c) The extended persistence diagram of the graph. In the diagram, each persistence point corresponds to the life span of the same color in (b).}
	\label{fig:PH}
	\vspace{-.15in}
\end{figure*}

This motivates us to enhance graph representations with advanced graph topological information, e.g., counting the number of connected components or loops. In the example of Figure~\ref{fig:motiv}, loop-counting separates the two graphs. Figure~\ref{fig:motiv}(a) contains two loops. Whereas, Figure~\ref{fig:motiv}(b) only has one.
To thoroughly exploit the topology of graphs, we resort to a robust loop-counting method based on the theory of topological data analysis \citep{edelsbrunner2010computational}. In particular, we use persistent homology \citep{edelsbrunner2000topological} to encode high-order topological features in a robust manner~\citep{cohen2007stability}. 
\emph{Persistent homology}, a modern development of the classic algebraic topology~\citep{munkres2018elements}, not only counts the number of topological structures, e.g., connected components and loops, but also measures their structural strength in view of a scalar function assigned to the graph, called the \emph{filter function}.
We filter a graph with the filter function and gradually add nodes and edges based on their function value. We capture topological structures that appear and disappear during the process. Such information, summarized as a 2D point set called a \emph{persistence diagram}, effectively captures structural information in a provably robust manner \citep{cohen2007stability}.
See Figure~\ref{fig:PH} for an illustration.

Persistent homology has been explored in graph learning tasks \citep{hofer2017deep, zhao2019learning, hofer2020graph, carriere2020perslay, horn2021topological, turkes2022effectiveness}. However, it was mainly used as a global summary of the whole graph. Such a summary is hard to be directly transferred into representations of individual nodes and edges. Having localized node/edge representations is crucial for tasks such as node classification and link prediction. Nodes/edges of very different local contexts should have different local representations, resulting in different class predictions.

We propose the first method to adapt persistent homology for localized graph representation. In particular, we make the following technical contributions. We focus on the neighborhood of individual nodes/edges (called \emph{vicinity graphs}), which were shown to carry rich local contextual information~\citep{zhang2018link, frasca2022understanding, qian2022ordered, huang2023boosting}. To capture the local connectivity information, we use distance functions weighted by different graph metrics as input filter functions for persistent homology computation. Theoretical results show that this localized topological feature encodes the connectivity richness around a node/edge of interest and achieves high expressive power. We focus on \emph{extended persistent homology (EPH)} \citep{cohen2009extending} instead of the classic persistent homology (PH). Unlike the classic PH, EPH captures the whole life span of a loop, and thus is much more suitable for graph data. 

These design choices empower us to acquire high quality localized topological features of nodes and edges. Furthermore,
we incorporate these features into GNNs for better prediction power.
Experiments show that the localized topological features greatly enhance the representation learning, and achieve state-of-the-art results on various node classification and link prediction benchmarks.

Our work builds upon two preliminary conference papers~\citep{zhao2020persistence,yan2021link}, which are the first to adapt persistent homology for localized node-level and edge-level representation learning. 
On top of these preliminary results, we add additional in-depth discussion and exploration regarding the expressiveness and learnability of localized persistence features. We hope these studies provide insights for further employment of persistence features. Below we summarize these new contributions.

\myparagraph{Expressiveness.} In terms of expressiveness, we discuss what connectivity information can be encoded by EPH, especially within the vicinity graphs. We provide examples that can be differentiated by EPH, but cannot be separated by even 3-WL or 4-WL\footnote{In summary, $k$-WL is an extension of 1-WL. It builds a coloring function of $k$-tuples, and tests the isomorphism between graphs by comparing the histograms of colors.}. 
In addition, we also prove that the proposed framework can distinguish most pairs of non-isomorphic graphs. This shows the richness of the topological information carried by localized topological features, and establishes the theoretical foundation of our proposed approach.

\myparagraph{Learnability: end-to-end learning of localized topological features.}
In recent years, it has been discovered that topological computation can be treated as a differentiable operator~\citep{gameiro2016continuation,poulenard2018topological}. This leads to novel ideas such as persistent-homology-based losses or regularizers~\citep{chen2019topological,hu2019topology,hofer2019connectivity}. 
In graph learning, different methods have been proposed to end-to-end learn topological features by backpropagating through the topological computation~\citep{hofer2020graph, carriere2020perslay}.
However, these methods are mainly focusing on learning global topological features. When it comes to localized topological features, the burden on computation can be very heavy.

In this paper, we thoroughly explore the pros and cons of end-to-end learning of local topological features. We provide two end-to-end learning strategies, corresponding to different levels of flexibility in feature generation. At the first level, we generate topological features based on different filter functions and learn to combine them through learning. The topological computation is still treated as a rigid feature generator.
At the second level, we do not use precomputed filter functions. Instead, we learn to directly generate a filter function and use its corresponding topological feature. On certain link prediction benchmarks, we observe improved performance of the end-to-end learning approaches. But the improvement is achieved at a much higher computational expense that can be prohibitive in practice. 

~

\smallskip In summary, our contribution is four-fold:

\begin{itemize}
    \item We propose a localized topological feature based on extended persistent homology. By focusing on vicinity graphs and using a distance-based filter function, our feature effectively captures the connectivity information of the local contexts of individual nodes and edges. 
    \item We integrate the localized topological feature into GNNs to enhance node/edge representation learning. Our methods achieve state-of-the-art node classification and link prediction results on various benchmarks. To the best of our knowledge, we are the first to use the persistent-homology-based feature to enhance node-level and edge-level representation learning.
	\item We extend the work from direct generation to end-to-end learning of localized topological features. We thoroughly evaluate the pros and cons empirically. The results and insights are valuable to the community for employing topological features.
	\item In terms of expressive power, we provide theoretical discussions of the structural information that can be encoded by localized persistent homology features, and show that it can effectively enhance graph learning models. We also propose a novel persistence feature based on our theoretical findings and previous works. This can serve as a justification for the proposed approach.
\end{itemize}

\myparagraph{Outline.} In Section~\ref{sec:related}, we briefly review existing graph representation learning and learning with topological information. In Section~\ref{sec:pre}, we introduce persistent homology and extended persistent homology. Section~\ref{sec:RL} covers the main contributions of this paper, including a theoretical discussion of the expressive power (Section~\ref{subsec:theory}), the proposed persistence-homology-enhanced representation learning GNNs (Section~\ref{subsec:PE}), as well as the end-to-end learning methods (Section~\ref{subsec:Learnable}). Finally, in Section~\ref{sec:exp}, we evaluate our framework on various node classification and link prediction benchmarks. We also provide detailed discussions on the end-to-end learning strategy (Section~\ref{subsec:exp_learn}).

\section{Related Works}
\label{sec:related}

\subsection{Representation learning on graphs}

Early representation learning methods~\citep{ng2002spectral, tenenbaum2000global, menon2011link} learn node representations that preserve the structural information of the input graph.
These methods are based on strong assumptions, and cannot generalize to various downstream tasks. Graph embedding methods~\citep{perozzi2014deepwalk, tang2015line, grover2016node2vec} map the structural information of an input graph into the embedding space using a skip-gram model.
However, these methods fail to utilize the intrinsic node features.

In recent years, GNN-based representation learning models have been proposed. Previous works~\citep{kipf2016semi, hamilton2017inductive, velivckovic2018graph, chami2019hyperbolic} use different mechanisms (e.g., attention~\citep{velivckovic2018graph} or hyperbolic convolution~\citep{chami2019hyperbolic}) to learn representations more effectively. 
However, without additional graph structural information, their expressive power in terms of differentiating non-isomorphic graphs is still limited~\citep{xu2018powerful}. 
For example, high-order graph convolutions have been proposed in topological graph representation learning
\citep{bodnar2021weisfeiler, bodnar2021weisfeiler2, bodnar2022neural, joshi2022expressive}.
One may also enhance the expressive power of GNNs with explicit structural features. An early example is GCN~\citep{kipf2016semi}, in which node degree is added as an additional node feature. One may also use statistics of relative distance from nodes of interest \citep{zhang2018link, you2019position} or other edge-level structural information~\citep{zhu2021neural, pan2022neural,huang2023link}. Our method continues along this direction. We propose to use localized topological features to encode the rich connectivity information. The expressive power of the new feature is validated both empirically and theoretically. 

\subsection{Learning with persistent homology}

Persistent homology~\citep{edelsbrunner2000topological, edelsbrunner2010computational} extracts high-level topological information of the input data based on the mathematics of algebraic topology~\citep{munkres2018elements}. The well-known stability result~\citep{cohen2007stability,chazal2009proximity} establishes the theoretical guarantee of the robustness of the extracted topological information against noise of measurements. This sets up the theoretical foundation of the application of persistent homology in modern data science.
In recent years, methods have been proposed to adapt persistent homology or related cycle information into machine learning settings, such as representation learning \citep{adams2017persistence, royer2021atol, hofer2017deep, yan2022cycle, yan2024cycle}, clustering \citep{ni2017composing}, kernel machines \citep{reininghaus2015stable, kusano2016persistence, carriere2017sliced}, property estimation~\citep{turkes2022effectiveness}, and topological loss~\citep{chen2019topological, hu2019topology, wang2020topogan, hofer2019connectivity, yan2022neural, stucki2023topologically}. 

For graph-structured data, persistent homology has been used for the graph classification task~\citep{zhao2019learning, hofer2020graph, carriere2020perslay, horn2021topological,immonen2023going}. 
In addition, some works focus on the end-to-end learning of topological features. These works either learn more effective topological features~\citep{zhao2019learning, hofer2020graph,carriere2020perslay}, or accelerate the inference of persistence homology features~\citep{de2022ripsnet, yan2022neural}. However, these methods mainly focus on graph-level representation, and cannot generate/learn topological features for individual nodes or edges. 
Our method is the first to concentrate on local context and generate/learn localized topological features for node/edge representation. 
We also note some following methods using localized topological features to restructure the graph for better representation learning~\citep{chen2021topological}. This method suffers from slow computational time in practice due to the expensive comparison between local topological features of pairs of nodes. 

\section{Preliminaries}
\label{sec:pre}

In this section, we introduce the preliminaries of our proposed framework: the theory of extended persistent homology, and refer the reader to~\citep{cohen2009extending,edelsbrunner2010computational} for more details.

\subsection{Ordinary Persistent Homology}

\textcolor{black}{In this paper, we focus mainly on finite simple graphs, where there are no loops or multiple edges. As defined in~\citep{munkres2018elements}, graphs are viewed as abstract simplicial complexes, consisting of a collection of finite, nonempty subsets of a vertex set $V$. It contains the singleton sets of $V$, and no subsets containing three or more elements. Specifically, the two-element subsets correspond to edges.}

In this context, persistent homology captures 0-dimensional (connected components) and 1-dimensional (loops) topological structures and measures their saliency via a scalar function called \emph{filter function}. 
Denote an input graph as $G = (V, E)$, where $V$ and $E$ represent the vertex and edge sets. We call all the nodes and edges \emph{simplices}. Denote by $X=V\cup E$ the set of all simplices. The filter function projects the simplex set to the real number set $f: X \rightarrow \mathbb{R}$. Often, the filter function is first defined on the vertex set: $f: V\rightarrow \mathbb{R}$, e.g., the node degree\footnote{\textcolor{black}{In this context, the filter function for nodes can be arbitrary.}}. Then it can be transferred to the edge set. For example, we can use $f(uv) = max(f(u), f(v))$, which will lead to a \emph{lower-star filtration} of the complex.

A \textit{sublevel set} of $X$ is defined as the subset of simplices whose filter function value is below or equal to a given threshold $t$, $X_t = \{x|f(x) \leq t, x \in X\}$. As $t$ increases from $-\infty$ to $\infty$, $X_t$ grows from an empty set to the whole complex. Such sequence of growing sublevel set is called an \textit{ascending filtration}: $\emptyset = X_{-\infty} = X_{t_0} \subset ... \subset X_{t_n} = X_{\infty} = X$. 
In Figure~\ref{fig:PH}(b), the first three denote its ascending filtration. In the first figure, the threshold is $t_1$, and $X_{t_1}$ only contains the vertex $u_1$. As $t$ increases to $t_2$ and $t_3$, more simplices are included in the ascending filtration, leading to the birth (appearance) or death (disappearance) of new topological structures. 
We use the \textit{persistence point} $(b, d)$ to measure the saliency of a topological structure, where $b$ and $d$ represent its birth and death time. The lifetime $|d-b|$ is then defined as its \textit{persistence} and intuitively measures its importance in terms of the input filter function. For instance, the green connected component in Figure~\ref{fig:PH}(b) is created (borns) when reaching $X_{t_2}$ and then combined with the whole connected component (dies) when reaching $X_{t_3}$, therefore its persistence point is $(t_2, t_3)$. 

\subsection{Extended Persistent Homology}

One shortcoming of ordinary persistent homology is it cannot capture the death of certain topological structures, e.g., loops and the whole connected component. These topological features are called \textit{essential features}. 
To represent the importance of these structures, Cohen-Steiner et al. \citep{cohen2009extending} introduce the \textit{extended persistence module}: $\emptyset = \homo(X_{-\infty}) \to \cdots \homo(X_a) \to \cdots \homo(X) = \homo(X, X^{\infty}) \to \cdots \to \homo(X, X^a) \to \cdots \to \homo(X, X^{-\infty})$, where \mbox{$X^a = \{x\in X|f(x) \geq a\}$} is a \emph{superlevel set} of $X$ at value $a$. The second part 
$\homo(X, X^{\infty}) \to \cdots \to \homo(X, X^a) \to \cdots \to \homo(X, X^{-\infty})$ is induced by a \textit{descending filtration}. 
According to~\citep{cohen2009extending}, the filter function of edges in the descending filtration is different from the previous definition: $f(uv) = min(f(u), f(v))$. This will lead to a \textit{upper-star filtration} of the complex.

An example is shown in the last three figures of Figure~\ref{fig:PH}(b). All loop features will be killed at the end of the extended sequence, and thus can be captured by certain persistence points whose birth happens in the ascending filtration. In general, the death time for loops will be smaller than their birth time. The number of such loops equals the 1st Betti number $\beta_1$ (rank of the 1st homology group).
For instance, the blue loop in Figure~\ref{fig:PH}(b) appears when reaching $X_{t_3}$ in the ascending filtration, and it is killed when entering $H(X, X^{t_1})$ in the descending filtration. Therefore, the persistence point of the blue loop is $(t_3, t_1)$.

Then we can define the birth-death times for all topological structures in the graph. Combining these persistence points (induced by both the standard ascending sequence and the extended sequence\footnote{This include persistence points with birth and death both from the ascending sequence, with birth and death both from the descending sequence, and with birth from the ascending, death from the descending sequence.}), we can obtain a planar multiset of points, called the \textit{extended persistence diagram} (EPD). An example is shown in Figure~\ref{fig:PH}(c). Notice that persistence diagrams are not specific to extended persistent homology, they can also be encoded with ordinary persistent homology\footnote{It can also be encoded using $0$th level set zigzag persistence.}. After obtaining the EPD, we can encode its information into a fixed-size vector called persistence image~\citep{adams2017persistence}.

\subsection{Persistence Image}
\label{subsec:PI}

EPDs live in an infinite-dimensional space, and cannot be directly used in machine learning frameworks. There have been many works to vectorize EPDs for downstream tasks, among which a popular choice is the \textit{persistence image}~\citep{adams2017persistence}. It can project the input EPD to a fix-sized vector. 

Denote $T:\mathbb{R}^2\rightarrow \mathbb{R}^2$ as a linear transformation for persistence points: $T(x,y) = (x,y-x)$. Given a EPD $D$, $T(D) = \{T(d)|d \in D\}$ represents the transformed diagram. For any $z \in \mathbf{R}^2$, $\phi_u(z)=\frac{1}{2\pi\sigma^2}e^{-\frac{||z-u||^2}{2\sigma^2}}$ is the 2D Gaussian function with mean $u$ and standard deviation $\sigma$. \textcolor{black}{In our setting, $\sigma$ is set to 1, and $u$ is set to 0.} $\alpha: \mathbb{R}^2 \rightarrow \mathbb{R}$ is a non-negative weight function. In our setting, it is a piece-wise linear weighting function\footnote{\textcolor{black}{The default setting of the Dionysus package~\citep{dionysus_package} reverses the birth and death values in the 1D EPD, ensuring that the new "death" value is greater than the "birth" value. In addition, we have normalized the filter function to a range between 0 and 1.}}:
\begin{equation}
\alpha(x,y)=
\begin{cases}
0 & \text{if } y \leq 0 \\
y & \text{if } 0 < y \leq 1 \\
1 & \text{if } y > 1
\end{cases}.
\end{equation}

Define the persistence surface of $D$ as $\rho_D(z) =  \sum_{u\in T(D)} \alpha(u)\phi_u(z)$. The persistence image is the collection of $n$ pixels $PI_D = \{PI_D[p]\} \in \mathbf{R}^n$ where every pixel is the integral of the persistence surface:  $PI_D[p] = \int \int_p \rho_{D}(x,y)dxdy$.
 \textcolor{black}{Persistence Image has been proven stable with respect to the $W_1$ of PD~\citep{adams2017persistence}. And it is known that PD is stable with respect to the $l_\infty$ norm of the filter function~\citep{edelsbrunner2010computational, cohen2007stability}. Since then \citet{skraba2020wasserstein, skraba2021notes} have shown the $W_1$ of the PD is stable with respect to $l_1$ norm of the filter function, if there is a fixed simplicial complex.}
Therefore the fixed-size vector can be widely adopted in machine learning models~\citep{carriere2020perslay, de2022ripsnet}.

\begin{figure*}[btp]
	\centering
	\subfigure[]{
		\begin{minipage}[t]{0.48\linewidth}
			\centering
			\includegraphics[width=\columnwidth]{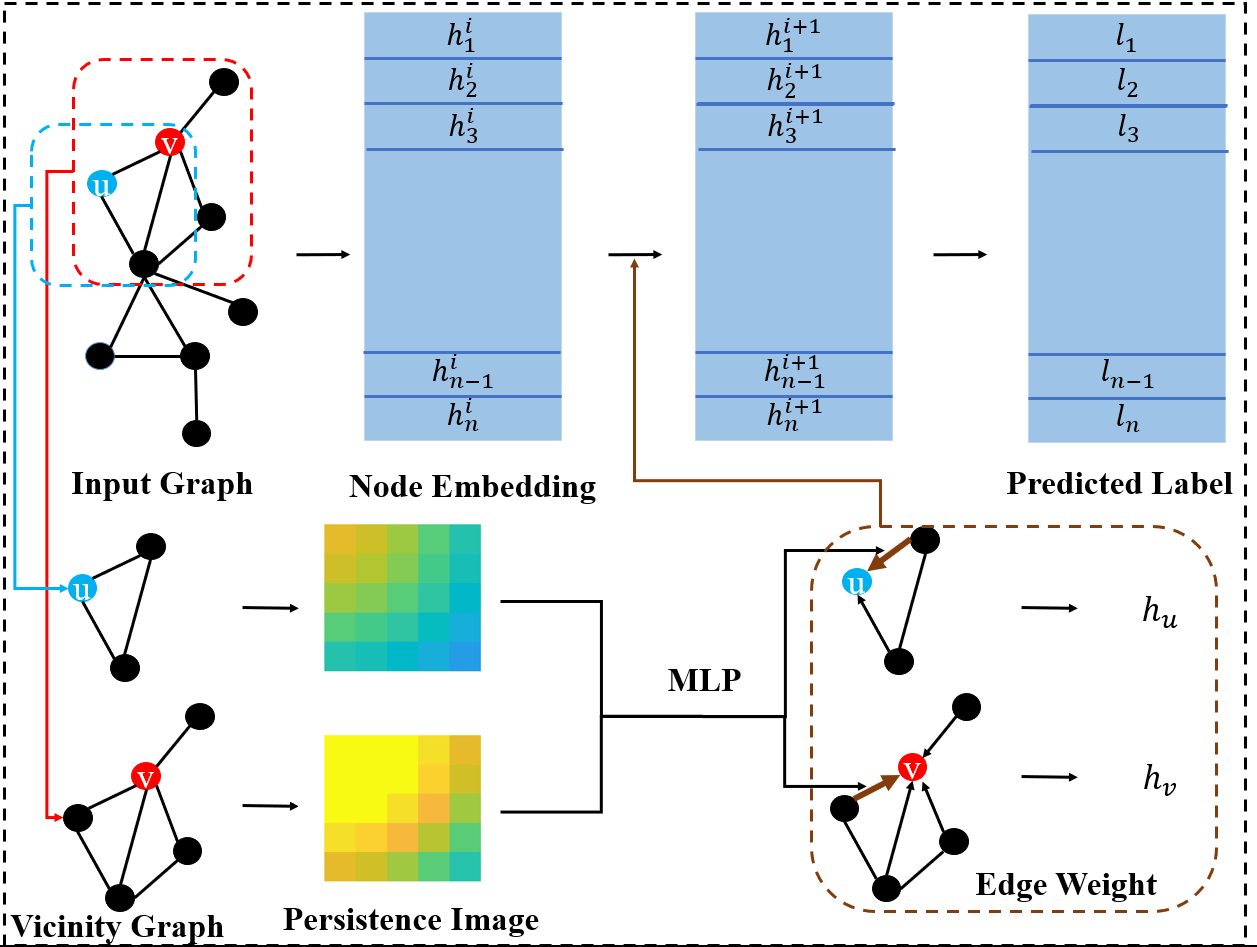}
		\end{minipage}%
	}%
	\subfigure[]{
		\begin{minipage}[t]{0.52\linewidth}
			\centering
			\includegraphics[width=\columnwidth]{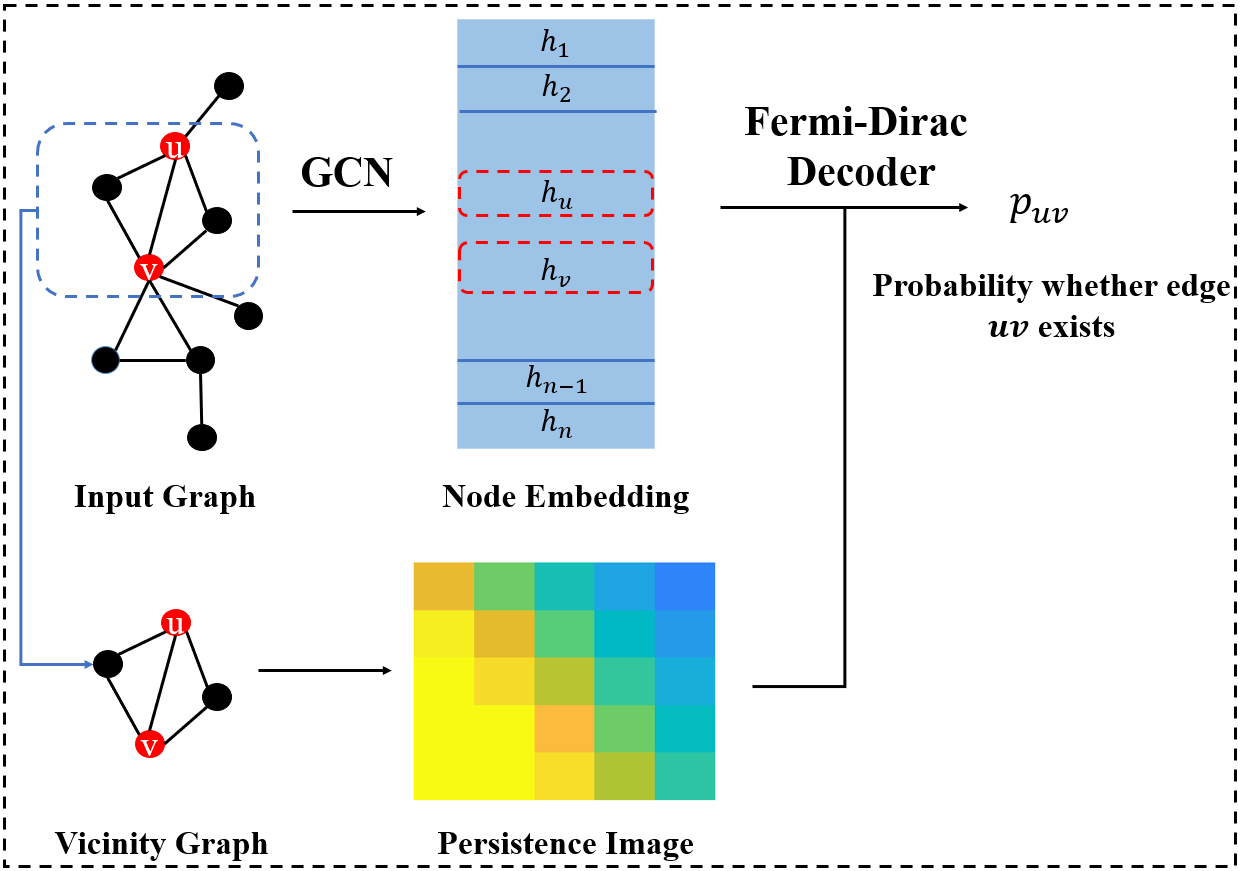}
		\end{minipage}
	}%
	\centering
	\vspace{-.1in}
	\caption{The architecture of \textcolor{black}{Persistence Enhanced Graph Network (PEGN)}. (a) The architecture for node classification. (b) The architecture for link prediction.}
	\label{fig:PEGN}
	\vspace{-.15in}
\end{figure*}

\section{Representation Learning with Extended Persistent Homology}
\label{sec:RL}

We have introduced how to obtain the topological features, i.e., EPDs, of an input graph. In this section, we present how to use such topological features in a localized manner to enhance the node/edge representations. In particular, we are extracting EPDs within the vicinity graph of each node/edge. 
Given a node $u$ in graph $G$, and let $G_u^k \subseteq G$ denote the $k$-hop neighborhood of $u$, that is, the subgraph around $u$ spanned by all nodes within $k$-hops to $u$. Formally speaking, $G_u^k = (V_u^k, E_u^k)$, where $V_u^k = \{v|d(u,v) \leq k\}$, $d(u,v)$ is the shortest path distance between the two nodes, and $E_u^k = E \cap (V_u^k)^2$ denotes edges whose two nodes both belong to $V_u^k$. Consider the shortest path distance to $u$, which is a function defined at nodes in $G^k_u$. Call those nodes at distance $j$ to $u$ as \emph{level-$j$ nodes}: $\{v|d(u,v) = j\}$. We then generate the EPDs and the persistence images for these vicinity graphs using the method (\textcolor{black}{the ascending filtration and descending filration}) described in Section~\ref{sec:pre}.

In this section, we present how the localized topological features can be incorporated into GNN learning.
In Section~\ref{subsec:theory}, we discuss the structural information that can be encoded by EPDs and present theoretical results on the expressiveness of persistence-enhanced models. We also propose an augmented persistence image encoding based on the theoretical results. In Section~\ref{subsec:PE}, we introduce our framework: Persistence Enhanced Graph Network (PEGN). In the framework, we extract the topological feature of nodes and edges, and use it to enhance the representation learning of the GNN model. In Section~\ref{subsec:Learnable}, we focus on learnability and introduce the end-to-end learning strategies for topological features. 

\subsection{Structural information encoded by EPDs}
\label{subsec:theory}

The EPD $D_u^k$ induced by this shortest path distance function encodes a rich family of structural information of $G^k_u$. For example, using the results of~\citep{TW19}, one can show that the following information can be easily recovered from $D_u^k$. 
\begin{enumerate}
\item The number of triangles incident to $u$, as well as the clustering coefficient at $u$.
\item The number of level-$j$ nodes for any $j\le k$, as well as the number of edges among level-$j$ nodes. 
\item The number of crossing edges from level-$j$ to level-($j+1$) for any $j<k$. 
\item The persistence diagram induced by the shortest distance function over the $j$-hop neighborhood $G_u^j$ around $u$, for any $j < k$. 
\item The shortest cycle incident to $u$. 
\end{enumerate}
Note that much of the above information will be hard for a standard message passing GNN to recover. 
In contrast, they can be recovered from EPDs by simple queries that can be easily implemented using Multi-Layer Perceptron (MLP). 
Using these results, it can be verified that if we add such EPDs as topological position embeddings to the original node feature, then the resulting enhanced GNN (1) is more expressive than 2-WL when we assume that the initial node features are node degree or constant (also proven in ~\citep{horn2021topological}); (2) is not less expressive than 3-WL (can differentiate some pair of graphs that 3-WL cannot differentiate) and is less expressive than 4-WL; (3) is not less expressive than 4-WL under certain assumptions; (4) can distinguish most pairs of non-isomorphic regular  graphs.
Therefore we adopt the distance-based filter functions in our frameworks since these functions have provable theoretical power. Below we state the theorem formally.

\begin{figure*}[btp]
	\centering
	\subfigure[]{
		\begin{minipage}[t]{0.43\linewidth}
			\centering
			\includegraphics[width=0.8\columnwidth]{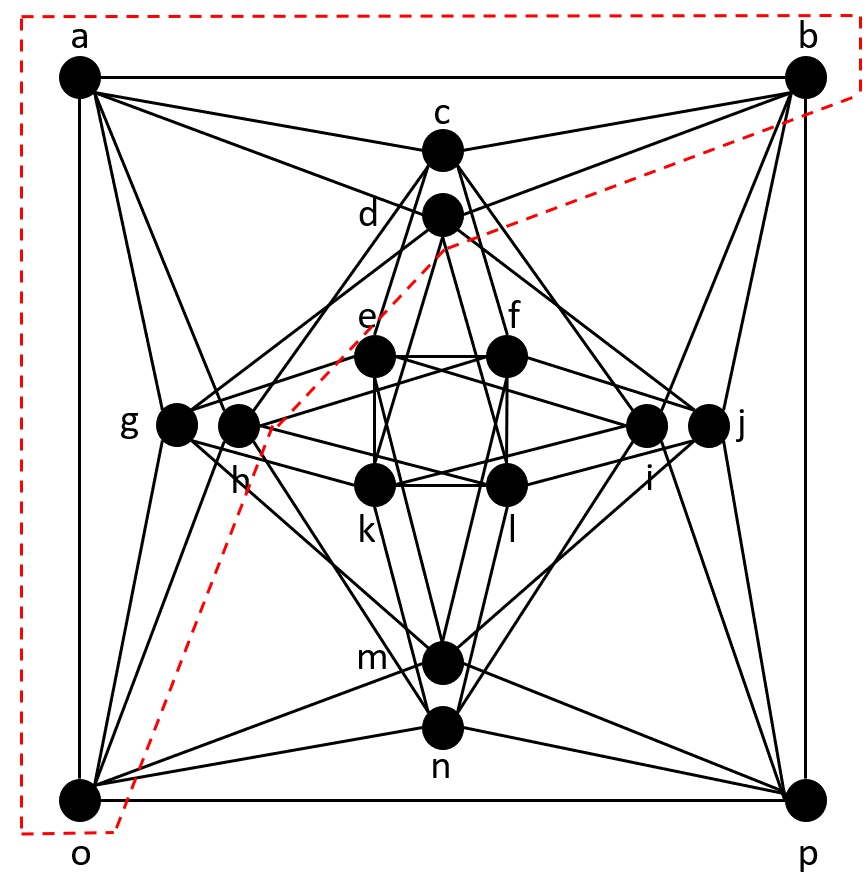}
		\end{minipage}%
	}%
	\subfigure[]{
		\begin{minipage}[t]{0.47\linewidth}
			\centering
			\includegraphics[width=0.8\columnwidth]{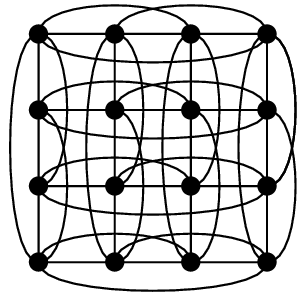}
		\end{minipage}
	}%
	\centering
	\vspace{-.1in}
	\caption{The pair of graphs that 3-WL cannot differentiate. (a) The Shrikhande Graph. (b) The 4*4 Rook Graph.}
	\label{fig:3WL}
	\vspace{-.15in}
\end{figure*}

\begin{theorem}
\label{th:ph_3wl}
With the shortest path distance function as the filter function, Extended Persistence Diagrams can differentiate some pairs of graphs that 3-WL cannot differentiate.
\end{theorem}

\begin{proof}
Consider the pair of graphs: the 4*4 Rook graph and the Shrikhande graph (shown in Figure~\ref{fig:3WL}). It has been shown that 3-WL cannot differentiate them~\citep{arvind2020weisfeiler}. Below we focus on the proof that EPDs can differentiate this pair of graphs.

We focus on the 1-hop neighborhood $G_a^1 = (V_a^1, E_a^1)$ of vertex $a$ in both two graphs (colored in red). In $G_a^1$, denote the filter function of all nodes $f_a: V \rightarrow R$ as the shortest path distance to $a$: $f_a(v) = d(v,a)$. We then extend the definition of the filter function to edges $f_a: E \rightarrow R$\footnote{\textcolor{black}{The function relaxation is applied to facilitate the proof. However, in the experimental section, we empirically observe that this relaxation does not affect the model's performance in Section~\ref{subsec:filter_discuss} in the Appendix. \textcolor{black}{Notice that the defined filter function used in the proof and the proof of Theorem~\ref{th:ph_4wl} is slightly different from the classic lower-star and upper-star filtration ~\citep{cohen2009extending}}.}}:  

\begin{equation}
\label{eq:filter}
f_a(uv)=
\begin{cases}
max(f_a(u), f_a(v)) & \text{if } f_a(u) \neq f_a(v) \\
f_a(u) + 0.5 & \text{if } f_a(u) = f_a(v)
\end{cases}.
\end{equation}

With the above definition, we can distinguish between the edges that connect $a$ and a layer-1 node, and edges that connect two layer-1 nodes. The filter value for the former edges is 1, and the filter value for the latter edges is 1.5. Therefore we can easily capture the 1-th Betti number (rank of the 1st homology group) of the induced subgraph $G_a$ formed by the layer-1 nodes of $a$. Formally speaking, the induced subgraph is $G_a = (V_a, E_a)$ where $V_a = \{u|d(u,a) = 1\}$, and $E_a = E \cap (V_a)^2$. In $G_a^1$, the number of 1D extended persistence points with persistence value \textcolor{black}{$(1.5, 0.5)$} denotes the 1-th Betti number of $G_a$: $\beta_1(G_a)$. Intuitively speaking, this is the number of loops formed by $G_a$. In the 4*4 Rook Graph, $G_a$ contains two 3-cycles: $bcd$ and $eim$, and $\beta_1(G_a) = 2$; while in the Shrikhande graph, $G_a$ contains one 6-cycle: $bchogd$, and $\beta_1(G_a) = 1$. Considering that the EPDs of these two graphs are different, we can differentiate the pair of graphs with EPDs.\footnote{This observation is made by Samantha Chen.} 

\textcolor{black}{In addition, the definition of descending filtration is provided below, and we can apply the same proof procedure as previously used.
\begin{equation}
\label{eq:filter_desc}
f_a(uv)=
\begin{cases}
min(f_a(u), f_a(v)) & \text{if } f_a(u) \neq f_a(v) \\
f_a(u) - 0.5 & \text{if } f_a(u) = f_a(v)
\end{cases}.
\end{equation}}
\end{proof}

We can extend the theorem whose proof is provided in the appendix:

\begin{theorem}
\label{th:4wl}
With the distance function~\citep{zhang2018link} as the filter function, Extended Persistence Diagrams can differentiate some pairs of graphs that 4-WL cannot differentiate.
\end{theorem}

We also present an upper bound on the expressiveness of 1D EPDs when the shortest path distance is used as the filter function. The proof is provided in the appendix.

\begin{theorem}
\label{th:ph_4wl}
With the shortest path distance as the filter function, the 1D Extended Persistence Diagram is less powerful than 4-WL in terms of differentiating non-isomorphic graphs.
\end{theorem}

In addition, although capable of distinguishing most pairs of graphs~\citep{babai1980random}, WL fails to distinguish many pairs of regular graphs. Therefore we provide the results on distinguishing regular graphs using EPDs. As shown by Theorem~\ref{th:ph_regular}, EPDs can differentiate most pairs of regular graphs. The proof is available in the appendix.

\begin{theorem}
\label{th:ph_regular}
Consider all pairs of $n$-sized $r$-regular graphs, let $ 3 \leq r < (2log2n)^{1/2}$ and $\epsilon$ be a fixed constant. With at most $K=\lfloor(\frac{1}{2}+\epsilon)\frac{\log{2n}}{\log{(r-1)}} \rfloor$, there exists an Extended Persistence Diagram with the shortest path distance as the filter function that distinguishes $1-o(n^{-1/2})$ such pairs of graphs.
\end{theorem}

\textbf{Persistence image plus.}  In Section~\ref{subsec:PI}, we vectorize the EPDs to persistence images since they are continuous fixed-size vectors that can be directly combined with GNNs. However, the mentioned valuable structural information, including the intra-layer and inter-layer connections, remains difficult to be extracted from persistence images, thus potentially diminishing the representation power. To address the limitation, we propose a structural-augmented persistence image formulation termed \textit{persistence image plus} (\textbf{PI+}): 
\begin{equation}
\label{eq:pi+}
PI+ = [PI;\{n_j|j \leq k\}; \{n_{j,j}|j\leq k\}; \{n_{j,(j+1)}|j < k\}],
\end{equation} where $[;]$ denotes the concatenation between vectors, \textcolor{black}{$PI$ is the persistence image derived from the concatenation of the 0D and 1D EPD,} $n_j$ denotes the number of level-$j$ nodes, $n_{j,j}$ denotes the number of edges among layer-$j$ nodes, and $n_{j,(j+1)}$ denotes the number of edges from level-$j$ to level-$(j+1)$. If $k$ is fixed, $PI+$ is also a fixed-size vector that can be directly combined with GNNs. In the experiment part, we will show the improved representation power of the proposed encoding.

\subsection{Persistence Enhanced Graph Networks}
\label{subsec:PE}
To capture the topological information of nodes and edges more effectively, we specially select the corresponding vicinity graph and filter function to generate persistence features according to different tasks. Below we explain our choices for node classification and link prediction \textcolor{black}{by first formulating these two tasks. Node classification and link prediction are two fundamental tasks in graph representation learning~\citep{zhang2020deep, wu2020comprehensive, ma2021deep}, each playing a crucial role in the analysis of graph-structured data. Given a graph \( G = (V, E, X) \), where \( V = \{1, \dots, n\} \) denotes the set of vertices, with \( n \) representing the number of nodes, \( E \) is the set of edges, and \( X \in \mathbb{R}^{n \times d} \) represents the node attributes (with the possibility of including edge attributes as well), where each row \( X_v \) corresponds to the \( d \)-dimensional feature vector of node \( v \in V \), these tasks leverage the graph's structure and features to make informed predictions.}

\subsubsection{Node Classification}

\textcolor{black}{\textbf{Task.} Given an input graph and the selected nodes in the graphs, we need to predict the labels of these nodes. 
The objective is to predict the labels or attributes of nodes based on the graph's structure and the features of the nodes. Specifically, given a label vector \( \mathcal{L} \in \mathbb{R}^{n} \), where each entry \( l_v \) represents the class label of node \( v \) for a subset of nodes \( V_{\text{train}} \subseteq V \), the goal is to predict the labels for an unlabeled node set \( V_{\text{test}} \subseteq V \setminus V_{\text{train}} \). The predicted label for each node \( v \in V_{\text{test}} \) is given by:
\[
\hat{l}_v = \text{argmax}_c \, P(l_v = c \mid G, X, \theta),
\]
where \( P(l_v = c \mid G, X, \theta) \) is the probability that node \( v \) belongs to class \( c \), given the graph structure \( G \), node features \( X \), and model parameters \( \theta \). This task is commonly applied in scenarios such as social network analysis, where the goal might be to determine a user's group affiliation or interests, or in biological networks, where it could involve predicting the function of proteins within a protein interaction network. }

In our framework, we capture the structural information of a given node $u$ by focusing on its $k$-hop neighborhood $G_u^k$. After obtaining the vicinity graph $G_u^k$, we need to define a filter function to compute its persistence image. To better capture the connectivity and structural information, we adopt a distance-based filter function following the inspiration of Section~\ref{subsec:theory}. To be specific, we first define a graph metric of $G_u^k$, then use the distance from node $v \in V_u^k$ to $u$ as its filter function. In our setting, we adopt Ollivier-Ricci curvature\footnote{\textcolor{black}{We use the GraphRicciCurvature package~\citep{GraphRicciCurvature} to compute. $\alpha$ is set to 0.5, and we adopt the approximated Sinkhorn Optimal Transportation Distance~\citep{NIPS2013_af21d0c9} for a faster computation.}} \citep{ni2018network}, which measures the distance taken to travel from neighbors of a node to neighbors of another node, as the graph metric. It is shown to be beneficial to graph convolutions both empirically and theoretically~\citep{ye2019curvature,topping2021understanding, li2022curvature}. To make all the edge weights positive, we add $1$ to Ollivier-Ricci curvature\footnote{\textcolor{black}{We observe that the curvature values computed in the adopted benchmarks are consistently greater than or equal to -1. The relevant data is available at \url{https://github.com/pkuyzy/curv_}.}}. Then we can obtain the persistence image $PI(u)$ for every given node $u$ in $G$ following the procedure of Section~\ref{sec:pre}. \textcolor{black}{We need to compute the EPDs for all nodes in the given graph.}

To effectively combine persistence images with our framework, we use them to reweight messages in GNNs. The proposed framework is shown in Figure~\ref{fig:PEGN}(a). The input of the model is the graph $G$, the original feature vectors of the nodes: $H^0 = [h_1^0, h_2^0,...,h_n^0] = X \in \mathbb{R}^{n*d}$, and the label set $\mathcal{L}$
of the dataset. Our task is to predict the label of the test nodes \( V_{\text{test}}\).

In our framework, we update the node representation by message passing and aggregation: 
\begin{equation}
h_u^k = \sigma_{k-1}(\sum_{v \in \overline{N}(u)} \alpha_{uv}^{k-1} W^{k-1}_1h_v^{k-1})
\end{equation}

Here, $h^k_u$ denotes the feature vector of node $u$ in the $k$-th layer, $\sigma_{k-1}$ denotes the activation function in the $(k-1)$-th layer, $\overline{N}(u) = \{v|d(u,v) \leq 1\}$ is the neighborhood of node $u$, $W^{k-1}_1$ is a learnable matrix to encode the feature vector in the $(k-1)$-th layer, and  $\alpha_{uv}^{k-1}$ represents the weight of edge $uv$ in the $(k-1)$-th layer. To be specific, the edge weight is computed by: 
\begin{equation}
\alpha_{uv}^{k-1} = \sigma(W^{k-1}_2[PI(u);PI(v)])
\end{equation}

In the formulation, $\sigma$ denotes the activation function, $W_2^{k-1}$ is a two-layer MLP, \textcolor{black}{which is a fully-connected neural network with a hidden layer size of 50 and a single output value,} and $[;]$ denotes the concatenation of the two persistence images. After the $L$-layer message passing, the output will be $H^L = [h_1^L, h_2^L, ..., h_n^L]$, and we use the cross-entropy loss between $H^L$ and the ground truth label $L$ to update our network.

\subsubsection{Link Prediction}
\label{subsubsec:LP}

\textcolor{black}{\textbf{Task.} The task focuses on inferring the existence of edges between pairs of nodes. Although the current structure of the graph is known, the objective is to predict unknown or potential future connections. Specifically, given a set of node pairs \( S = \{(u, v) \mid u, v \in V \} \), where each pair is either connected by an edge \( (u, v) \in E \) (positive samples) or not \( (u, v) \notin E \) (negative samples), the goal is to predict whether there is a potentially true link between the given pair of nodes \( (u, v) \in S \). The prediction is formulated as:
\[
\hat{Y}_{uv} = \text{activation}(\phi(u, v, G, X, \theta)),
\]
where \( \phi(u, v, G, X, \theta) \) is a scoring function that computes the likelihood of an edge existing between nodes \( u \) and \( v \), given the graph structure \( G \), node features \( X \), and model parameters \( \theta \). The activation function ensures the output is a probability between 0 and 1. Link prediction is particularly useful in scenarios where the graph is incomplete or evolving, such as in social networks for suggesting new connections or in recommender systems for identifying potential associations between users and products.} \textcolor{black}{As illustrated later, the activation function is the Fermi-Dirac decoder~\citep{krioukov2010hyperbolic, nickel2017poincare}, and $\phi$ is the proposed GNN.} 

Previous works~\citep{zhang2021labeling,zhang2018link} show that the relationship between the target nodes contributes to predicting missing links. Therefore, we focus on modeling the interaction between the target nodes rather than the \textit{node-wise} information~\citep{chami2019hyperbolic}. In the later section, we will use experiments to show the superiority of our proposed \textit{pair-wise} topological representation for link prediction.

The architecture of our framework is shown in Figure~\ref{fig:PEGN}(b), the input of the model is the whole graph and multiple pairs of target nodes, and the output is the probability of whether edges exist between these pairs of nodes. The proposed framework consists of two branches: the upper branch adopts a graph neural network to extract the node embeddings of the input graph, and the lower branch computes the persistence image of the vicinity graph of these node pairs. 

To focus on the interaction between a pair of target nodes, $u$ and $v$, we adopt the intersection of the $k$-hop neighborhoods of $u$ and $v$ as the vicinity graph. To be specific, let $G = (V,E)$ be the input graph. $V_u^k$ and $V_v^k$ are the $k$-hop neighborhoods of target nodes $u$ and $v$. $V_{uv}^k = V_u^k \cap V_v^k$ is the intersection of the two neighborhoods. The vicinity graph is then defined as $G_{uv}^k = (V_{uv}^k, E_{uv}^k)$, where $E_{uv}^k = E \cap (V_{uv}^k)^2$, denoting that when two nodes are connected in the original graph, they will also be connected in the vicinity graph. Notice here the vicinity graphs are selected based on the hop distance. In the later section, the filter function will be defined based on certain graph metrics.

Similar to the distance-based function for node classification, we define the filter function of a node $i\in V_{uv}^k$ as the sum of its distances to the target nodes based on a specific graph metric: $f(i) = d_o(i,u) + d_o(i,v)$. Again, we adopt Ollivier-Ricci curvature as the metric function. Then we can compute the persistence image for the vicinity graph, denoted as $PI(u, v)$.

In the upper branch of our framework, we obtain the node embeddings through a classic $L$-layer graph convolutional network (GCN)\footnote{\textcolor{black}{In our setting, $L$ is set to 2.}} \citep{kipf2016semi}. After the $L$-layer GCN, the node representations are aggregated from their $L$-hop neighborhoods. Denote the final output node embedding as $H = [h_1, ..., h_{n}]$, where $n$ is the number of nodes, $h_i \in \mathbb{R}^d$ is the output representation of node $i$, $i = 1,2,...,n$.

Previous works~\citep{chami2019hyperbolic} learn representations based on the non-Euclidean attribute of graph-structured data. Inspired by their success, we adopt a modified Fermi-Dirac decoder \citep{krioukov2010hyperbolic, nickel2017poincare} to combine the representations from the upper branch and the lower branch. To be specific, we concatenate the distinction between the embeddings of target nodes $(h_u - h_v)$ with the persistence image $PI(u,v)$, and pass it to a two-layer MLP, which outputs a single value $FD(u,v) = MLP((h_u - h_v);PI(u,v))$. The probability of whether $uv$ exists is $prob(uv) = 1/(e^{(FD(u,v)-2)} + 1)$. We train the proposed framework by minimizing the cross-entropy loss with negative sampling\citep{zhang2018link,you2019position}.

\subsection{End-to-end learning of topological features}
\label{subsec:Learnable}

In recent years, it has been discovered that topological computation can be treated as a differentiable operator~\citep{gameiro2016continuation,poulenard2018topological}. In graph learning, different methods have been proposed to end-to-end learn topological features by backpropagating through the topological computation~\citep{hofer2020graph, carriere2020perslay}. 
However, these methods are mainly focusing on learning global topological features.

In this section, we focus on localized learnable topological features, and try to enhance the node representation and edge representation with end-to-end learning strategies.

\subsubsection{Combination of topological features}
\label{subsubsec:agg}

Given a set of persistence diagrams $\mathcal{D}$, previous works use approaches such as the expected persistence diagram~\citep{chazal2018density} to build a statistical summary of these diagrams. Following the inspiration of these works, for a given graph $G$, we generate topological features based on different filter functions and learn to combine them through learning. 

To obtain better topological representation, we carefully select filter functions that contain beneficial information of $G$. Following~\citep{carriere2020perslay}, we introduce a spectral signature named the Heat Kernel Signature (HKS)~\citep{hu2014stable, sun2009concise} as the filter function to fully exploit the informative feature of the input graphs. To be specific, let $A$ and $D$ be the adjacency matrix and degree matrix of the graph\footnote{$D$ is a diagonal matrix with $D_{i,i} = \sum_j A_{i,j}$.}. Then the normalized laplacian matrix of $G$ is $L = I - D^{-\frac{1}{2}}AD^{-\frac{1}{2}}$. Denote $\Psi = \{\psi_1, \psi_2,...\psi_n\}$ as the eigenvectors of $L$ whose corresponding eigenvalues satisfy $0 \leq \lambda_1 \leq \lambda_2 \leq ... $. Given a diffusion parameter $t$, the HKS is defined as: hks$_{G,t} = \sum_{k=1}^n exp(-t\lambda_k)\psi_k^2 \in \mathbb{R}^n$, where $n$ is the number of vertices in $G$. Therefore, we can assign every node a specific filter value.

In our settings, for every vicinity graph, we compute the HKS and use it as the filter function. Following \citep{carriere2020perslay}, the diffusion parameter $t$ of HKS is set to 0.1 or 10. Equipped with the three filter functions (Ollivier-Ricci curvature and two HKS functions), we can generate three persistence images for each vicinity graph. After that, we replace the original persistence image with the concatenation of these persistence images in our proposed frameworks. 

\subsubsection{Learnable filter function}
\label{subsubsec:filter}
Another approach is to generate a filter function rather than using precomputed filter functions. Hofer et al. \citep{hofer2020graph} have shown that gradient passing is possible for the computation of PDs, providing the basis for our method. 

In our setting, we initialize the filter function of every node in the whole graph as its node degree and pass the whole graph to several graph convolutional blocks~\citep{xu2018powerful} to learn the filter values. After that, we compute the persistence images of all vicinity graphs. The remaining procedures are the same as mentioned in Section~\ref{subsec:PE}.

In our original frameworks, to decrease the computational cost, we pre-compute the persistence images for all possible vicinity graphs at the beginning to avoid computing the persistence images in every epoch. However, when applying learnable filter function, the filter values of simplices are constantly updated, thus the persistence images should be updated in every epoch. To deal with the heavy computational burden, we adopt several approaches: (1)
we compute filter values for all nodes in the whole graph once, rather than compute different filter values in different vicinity graphs; (2) we compute 0D ordinary PDs rather than 1D EPDs. The computation can be done using a Union-Find algorithm~\citep{edelsbrunner2010computational}, with complexity $O(|E|\alpha(|E|))$~\citep{cormen2022introduction}, where $\alpha(.)$ is the inverse Ackermann function. While in practice, the complexity of computing EPDs is $O(|E|log|V|)$~\citep{zhang2022gefl} using the data structure of mergeable trees~\citep{georgiadis2011data, agarwal2006extreme}.

\section{Experiments}
\label{sec:exp}

To evaluate the effectiveness of our proposed frameworks, we compare them with various state-of-the-art methods on two graph representation learning tasks: node classification and link prediction. 
In Section~\ref{subsec:NC} and Section~\ref{subsec:LP}, we evaluate our framework on various node classification and link prediction benchmarks, respectively. In Section~\ref{subsec:exp_learn}, we discuss the pros and cons of the end-to-end learning strategies that generate topological features. Source code is available at \href{https://github.com/pkuyzy/TLC-GNN}{https://github.com/pkuyzy/TLC-GNN}.

\textbf{Datasets.} 
The dataset information is summarized below:
\begin{enumerate}
    \item \textbf{Cora, Citeseer, and PubMed} \citep{sen2008collective} are standard citation networks where nodes represent scientific papers, and edges denote citations between them. \textcolor{black}{The classes correspond to different topics or subjects, such as research fields or disease categories. The features for each node (which represents a paper) are typically binary vectors indicating the presence or absence of specific words (Cora and Citeseer), or TF-IDF values of words (PubMed).}
    \item \textbf{Photo and Computers} \citep{shchur2018pitfalls} are graphs derived from Amazon shopping records. In these graphs, nodes represent products, and an edge between two nodes indicates that the products are frequently bought together. \textcolor{black}{The classes represent different product categories, such as electronics, home appliances, or clothing. The features are derived from product metadata.}
    \item \textbf{Physics and CS} \citep{shchur2018pitfalls} are co-authorship graph datasets where nodes represent authors, and an edge between two nodes indicates that they have co-authored a paper. \textcolor{black}{The classes correspond to various research fields or domains. The features are derived from the authors' publication histories.}
    \item \textbf{PPI Networks} \citep{zitnik2017predicting} are protein-protein interaction networks originally designed for graph classification tasks. In this dataset, nodes represent proteins, and edges denote interactions between them. \textcolor{black}{There are no node-level classes, and the features are derived from various biological attributes of the proteins.} Each graph contains 3,000 nodes with an average degree of 28.8, and the dimensionality of the node feature vectors is 50.
\end{enumerate}

We list the details of these datasets in Table~\ref{tab:data}. Since PPI contains 20 different graphs, we do not list the details of all these graphs in the table. \textcolor{black}{As shown in the table, Cora and Citeseer are relatively small and sparse graphs, whereas the other datasets are larger and, in some cases, denser (e.g., Photo and Computers).} Some of the datasets consist of more than 200,000 edges, making representation learning more challenging.

\begin{table*}[t]
	\caption{Statistics of benchmark datasets}
	\label{tab:data}
	\begin{center}
		\begin{small}
			\begin{sc}
				\scalebox{1.0}{
					\begin{tabular}{lccccc}
						\hline\noalign{\smallskip}
						Dataset & Classes & Features & Nodes & Edges & Edge density \\
						\noalign{\smallskip}\hline\noalign{\smallskip}
						Cora & 7 & 1433 & 2708 & 5429 & 0.0015 \\
						Citeseer & 6 & 3703 & 3327 & 4732 & 0.0009\\
						PubMed  & 3 & 500 &  19717 & 44338 &  0.0002  \\
						Photo & 8 & 745 & 7487 & 119043 & 0.0042 \\
						Computers & 10 & 767 & 13381 & 245779 & 0.0027 \\
						Physics & 5 & 8415 & 34493 & 282455 & 0.0005 \\
						CS & 15 & 6805 & 18333 & 100227 & 0.0006\\
						
						\noalign{\smallskip}\hline
					\end{tabular}
				}
			\end{sc}
		\end{small}
	\end{center}
	\vskip -0.1in
\end{table*}

\subsection{Node classification}
\label{subsec:NC}

\begin{table*}
	\centering
	\caption{Node classification accuracy on various benchmark datasets}
	\label{tab:NC}
	\scalebox{0.7}{
		\begin{tabular}{lccccccc}
			\hline\noalign{\smallskip}
			Method & Cora & Citeseer &  PubMed & CS & Physics & Computers & Photo \\
			\noalign{\smallskip}\hline\noalign{\smallskip}
			MoNet \citep{monti2017geometric} & 81.7$\pm$0.0 & 71.2$\pm$0.0 & 78.6$\pm$2.3 & 90.8$\pm$0.6 & 92.5$\pm$0.9 & 83.5$\pm$2.2 & 91.2$\pm$1.3 \\
			GraphSAGE \citep{hamilton2017inductive} & 79.2$\pm$0.0 & 71.2$\pm$0.0 & 77.4 $\pm$ 2.2 & 91.3 $\pm$2.8 & 93.0$\pm$0.8 & 82.4$\pm$ 1.8 & 91.4 $\pm$1.3\\
			Graph U-Net \citep{gao2019graph} & 82.5$\pm$0.0 & 72.0$\pm$0.0 & 78.9$\pm$0.0 & 92.7$\pm$0.0 & 94.0 $\pm$0.0 & 86.0$\pm$0.0 & 91.9$\pm$0.0 \\
			WLCN \citep{morris2019weisfeiler} & 78.9$\pm$0.0 & 67.4$\pm$0.0 & 78.1$\pm$0.0 & 89.1$\pm$0.0 & 90.7$\pm$0.0 & 67.6$\pm$0.0 & 82.1$\pm$0.0 \\
			GCN \citep{kipf2016semi} & 81.5$\pm$0.5 & 70.9$\pm$0.5 & 79.0$\pm$0.3 & 91.1$\pm$0.5 & 92.8$\pm$1.0 & 82.6$\pm$2.4 & 91.2$\pm$1.2 \\
			GAT \citep{velivckovic2018graph} & 83.0$\pm$0.7 & 72.5$\pm$0.7 & 79.0$\pm$0.3 & 90.5$\pm$0.6 & 92.5$\pm$0.9 & 78.0$\pm$19.0 & 85.1$\pm$20.3 \\
			HGCN \citep{chami2019hyperbolic} & 78.0$\pm$1.0 & 68.0$\pm$0.6 & 76.5$\pm$0.6 & 90.5 $\pm$ 0.0& 91.3$\pm$0.0 & 82.1$\pm$0.0 & 90.5$\pm$0.0\\
            ARMA~\citep{bianchi2021graph} & 82.3$\pm$0.5 & 71.7$\pm$ 0.6 & 78.3$\pm$0.8 & 93.0$\pm$ 0.5 & 92.9$\pm$0.6 & 81.1$\pm$1.1 & 91.7$\pm$0.8\\
            CGNN-EXP~\citep{li2022curvature} & 82.5$\pm$0.6 & 72.1$\pm$0.7 & 78.9$\pm$0.5 & 93.2$\pm$0.3 & 93.7$\pm$0.3 & 83.8$\pm$0.7 & 91.4$\pm$0.6\\
			\noalign{\smallskip}
			\hline
			\noalign{\smallskip}
			PEGN  & 82.7$\pm$0.4 & 71.9$\pm$0.5 & \textbf{79.4$\pm$0.7} & \textbf{93.3$\pm$0.3} & 94.3$\pm$0.1 & 86.6$\pm$0.6 & 92.7$\pm$0.4\\
			
			GCN(+PI)  & 82.9$\pm$0.6 & \textbf{72.6$\pm$0.7} & 78.6$\pm$0.4 & 92.3$\pm$0.3 & 93.9$\pm$0.3 & 86.7$\pm$0.1 & 92.5$\pm$0.6\\
			GCN(+PI+) & \textbf{83.6$\pm$1.2} & \textbf{72.6$\pm$0.7} & 78.8$\pm$0.5 & 93.0$\pm$0.3 & 94.1$\pm$0.3& \textbf{86.9$\pm$0.5} &\textbf{93.0$\pm$0.7}\\
        \noalign{\smallskip}
			\hline
			\noalign{\smallskip}
			
			PEGN (FF) & 81.8 $\pm$ 0.5 & 71.8 $\pm$ 0.3& 78.7$\pm$0.4 & 92.7$\pm$0.4 & \textbf{94.4$\pm$0.2} & 86.2$\pm$0.2 & 92.6$\pm$0.5   \\
			PEGN (PI) & 81.4$\pm$0.7 & 71.3$\pm$ 0.9 & 78.3$\pm$0.5 & 93.0$\pm$0.4 & 94.2$\pm$0.4 & 84.9$\pm$1.1 & 90.2$\pm$0.7\\
			\noalign{\smallskip}\hline
	\end{tabular}}
\end{table*}
\textbf{Baselines.} For node classification, we compare with various baselines which adopt different approaches to enhance the message passing of GNNs. The baselines include \textbf{GCN} \citep{kipf2016semi} which reweights messages with node degrees,\textbf{ GraphSAGE} \citep{hamilton2017inductive} which aggregates the information from a sampled neighborhood with mean pooling, \textbf{MoNet} \citep{monti2017geometric}, \textbf{GAT} \citep{velivckovic2018graph}, \textbf{Graph U-Net} \citep{gao2019graph} which take advantage of self-attention mechanism in message passing, \textbf{WLCN} \citep{morris2019weisfeiler} which incorporates the information of subgraph structures in representation learning, \textbf{HGCN} \citep{chami2019hyperbolic} which leverages the non-euclidean attribute of the input graphs for representation learning, \textbf{ARMA}~\citep{bianchi2021graph} which is based on the spectral graph theory, and \textbf{CGNN}~\citep{li2022curvature}, which also incorporates additional topology-based feature to GNN models. 

\textbf{Experimental setup.} We use Cora, Citeseer, PubMed, CS, Physics, Computers, and Photo as the benchmarks. Considering the different densities and scales of these datasets, we focus on the 2-hop neighborhoods for Cora, Citeseer, PubMed, and CS, and 1-hop neighborhoods for Physics, Computers, and Photo. The dimension of persistence images is 25. We split the training, validation, and test set following~\citep{kipf2016semi, velivckovic2018graph}. To be specific, the training set consists of 20 nodes from each class, and the validation (test, resp.) set consists of 500 (1000, resp.) nodes. We initialize the model with Glorot initialization and use cross-entropy loss and Adam optimizer to train our model. In the optimizer, the learning rate is 0.005, and the weight decay is 0.0005. The training epoch is set to 200, and the early stopping on the validation set is 100 (patience epochs). For a fair comparison, we set the number of node embeddings of the hidden layer to be the same (64) for all networks. For all the models, the number of GNN layers is set to 2. All the activation function after each graph convolution block is ELU. Following \citep{kipf2016semi, velivckovic2018graph}, accuracy is used as the evaluation metric. \textcolor{black}{We utilize the Python package Dionysus, version 2.0.7, to compute the EPD. The subsequent experiments follow the same approach.}

\textbf{Results.} We present the results in Table~\ref{tab:NC}. In the table, \textbf{PEGN} denotes the proposed model, \textbf{GCN(+PI)} and \textbf{GCN(+PI+)} represent frameworks in which we concatenate the persistence image (or the PI+ from Equation~(\ref{eq:pi+}), resp.) with the initial node feature, and use GCN as the backbone. As illustrated in Section~\ref{subsec:theory}, the direct combination of the intrinsic node attributes and the additional topological features leads to better expressiveness. In the table, PEGN performs comparably with baseline methods on Cora and Citeseer, and outperforms baseline methods on the rest of the benchmarks. This demonstrates that the topological information captured by persistent homology enhances the representation learning of GNNs. While the relatively less improvement on small and sparse datasets, Cora and Citeseer, can be caused by their limited topological information. In these situations, the integrated topological feature cannot provide beneficial information. We also note that GCN(+PI+) outperforms baseline methods and GCN(+PI) on most benchmarks, showing that the proposed representation, PI+, preserves more essential topological information compared to the original persistence image.

\subsection{Link Prediction}
\label{subsec:LP}
\begin{table*}
	\centering
	\caption{AUC-ROC results on real-world link prediction datasets}
	\label{tab:LP} 
	\scalebox{0.9}{
		\begin{tabular}{lccccc}
			\hline\noalign{\smallskip}
			Method & Cora & Citeseer & PubMed & Photo & Computers \\
			\noalign{\smallskip}\hline\noalign{\smallskip}
			GCN \citep{kipf2016semi} & 90.5$\pm$ 0.2 & 82.6$\pm$1.9 & 89.6$\pm$3.7 & 91.8$\pm$0.0 & 87.8$\pm$0.0\\
			GAT \citep{velivckovic2018graph} & 72.8$\pm$ 0.2 & 74.8$\pm$1.5 & 80.3$\pm$0.0&  92.9$\pm$0.3& 86.4$\pm$0.0\\
			HGCN \citep{chami2019hyperbolic} & 93.8$\pm$0.1  & \textbf{96.6$\pm$0.1} & 96.3$\pm$0.0* & 95.4$\pm$0.0 & 93.6$\pm$0.0  \\
			P-GNN \citep{you2019position} &  74.1$\pm$2.4 & 73.9$\pm$2.6 &  79.6$\pm$0.5 & 90.9$\pm$0.7 & 88.3$\pm$1.0\\
			SEAL \citep{zhang2018link} & 91.3$\pm$5.7 & 89.8$\pm$2.3& 92.4$\pm$1.2 & 97.8$\pm$1.3 & 96.8$\pm$1.5 \\
    NBFNet~\citep{zhu2021neural} & 86.1$\pm$0.6 & 85.1$\pm$0.3 & OOM & 98.3$\pm$0.4 & OOM \\
    WalkPool~\citep{pan2022neural} & 92.3$\pm$0.3 & 90.7$\pm$0.4 & OOM & OOM & OOM \\
    Gelato~\citep{huang2023link} & 83.1$\pm$0.1 & 88.4$\pm$0.0 & 64.9$\pm$0.1 & 98.0$\pm$0.0 & 96.7$\pm$0.0 \\
			\noalign{\smallskip}
			\hline
			\noalign{\smallskip}
			PEGN (NC)  & 93.0$\pm$0.6 &  93.1$\pm$0.7 &  95.8$\pm$0.1  & 96.9$\pm$0.1 & 96.0$\pm$0.1\\
			PEGN (node-wise)  & 94.0$\pm$ 0.3 & 92.9$\pm$0.6 & 96.9$\pm$0.2 & 97.9$\pm$0.1 & 97.0$\pm$0.1\\
			\noalign{\smallskip}
			\hline
			\noalign{\smallskip}
			PEGN & 94.9$\pm$0.4 & 95.1$\pm$ 0.7 & \textbf{97.0$\pm$0.1} & 98.2$\pm$0.1 & 97.9$\pm$0.1\\
			\noalign{\smallskip}\hline\noalign{\smallskip}
			PEGN(PI) & \textbf{95.0$\pm$0.3} & 93.1$\pm$0.7 & 96.8$\pm$0.2 & \textbf{98.4$\pm$0.1} & \textbf{98.3$\pm$0.1}\\
			\hline
	\end{tabular}}
        \vskip -0.15 in
\end{table*}

\begin{table*}
	\centering
	\caption{AUC-ROC results on PPI link prediction datasets}
	\label{tab:PPI}
	\scalebox{0.75}{
		\begin{tabular}{lcccccccccc}
			\hline\noalign{\smallskip}
			Method & 1 & 2 & 3 & 4 & 5 & 6 & 7 & 8 & 9 & 10\\
			\noalign{\smallskip}\hline\noalign{\smallskip}
			GCN \citep{kipf2016semi} & 75.83 & 74.16 & 75.21 & 75.68 & 74.42 & 68.24 & 75.26 & 70.65 & 68.52 & 77.68 \\
			GAT \citep{velivckovic2018graph} & 57.55 & 51.81&  72.45 & 49.76 & 55.68& 47.03&  51.66 & 70.93 & 51.62& 50.69\\
			SEAL \citep{zhang2018link} & \textbf{85.82}  & 82.25 & 50.00 & 83.69 & 64.78 & 80.65 & \textbf{85.34} & 79.48 & 80.13 & 67.13\\
			P-GNN \citep{you2019position} & 80.20 & 81.62&75.33 & 78.89& 76.83& 80.64& 80.30& 82.21& \textbf{82.12} & 81.30\\
			\noalign{\smallskip}\hline\noalign{\smallskip}
			PEGN & 83.71 & 80.69 & 83.69 & 83.67 & 81.13 & 80.40 & 80.69 & 82.83 & 80.14 & 83.86 \\
			\noalign{\smallskip}\hline\noalign{\smallskip}
			PEGN (FF) & 85.47 & \textbf{83.14} & \textbf{84.32} & \textbf{85.27} & \textbf{83.88}& \textbf{81.91} & 84.30 & \textbf{85.63} & 81.96 & 84.00 \\
			PEGN (PI) & 83.98 & 81.17 & 83.40 & 82.94 & 82.11 & 78.26 & 80.74 & 83.54 & 78.91 & \textbf{84.33} \\
			\noalign{\smallskip}\hline
	\end{tabular}}
	\scalebox{0.75}{
		\begin{tabular}{lcccccccccc}
			\hline\noalign{\smallskip}
			Method & 11 & 12 & 13 & 14 & 15 & 16 & 17 & 18 & 19 & 20 \\
			\noalign{\smallskip}\hline\noalign{\smallskip}
			GCN \citep{kipf2016semi} & 76.21 & 74.90 & 74.26 & 77.91 & 69.66 & 75.40 & 76.68 & 77.36 & 78.02 & 77.40 \\
			GAT \citep{velivckovic2018graph} & 53.76& 50.70 & 47.94& 45.26& 66.09& 71.18& 42.15& 75.06& 52.74& 61.52\\
			SEAL \citep{zhang2018link} & 72.55  & 81.69 & 81.68 & 83.62 & 50.00 & \textbf{82.74} & 80.05 & 82.32 & 71.16 & 83.44\\
			P-GNN \citep{you2019position} & 73.77& 81.36& 73.93& \textbf{84.48}& 81.58& 82.09& 79.42& 82.37& 81.67& 81.70\\
			\noalign{\smallskip}\hline\noalign{\smallskip}
			PEGN  & 82.02 & 81.92 & 81.88 & 84.30 & 84.30 & 81.43 & 83.58 & \textbf{83.90} & 84.13 & 84.69\\
			\noalign{\smallskip}\hline\noalign{\smallskip}
			PEGN (PI) & \textbf{83.13} & \textbf{82.53} & \textbf{81.92} & 84.36 & \textbf{84.75} & 82.11 & \textbf{84.63} & 83.59 & \textbf{84.43} & \textbf{85.38}\\
			\noalign{\smallskip}\hline
	\end{tabular}}
        \vskip -0.2in
\end{table*}

\textbf{Baselines.} For link prediction, we compare with diverse baselines including popular link prediction methods such as \textbf{SEAL} \citep{zhang2018link}, \textbf{P-GNN} \citep{you2019position}, \textbf{NBFNet}~\citep{zhu2021neural}, \textbf{WalkPool}~\citep{pan2022neural}, and \textbf{Gelato}~\citep{huang2023link}, which augment machine learning models with edge-level structural information. We also compare with some general representation learning models like \textbf{GCN}, \textbf{GAT}, and \textbf{HGCN}. Notice that GCN and GAT are not directly used for link prediction. Therefore in our settings, we obtain node embeddings using these models and then adopt the Fermi-Dirac decoder~\citep{krioukov2010hyperbolic, nickel2017poincare} to predict the probability of whether edges exist.  

In addition, to demonstrate the importance of designing the \textit{pair-wise} feature rather than the \textit{node-wise} feature (proposed for node classification), we compare with our framework for node classification, which is named \textbf{PEGN (NC)} in Table~\ref{tab:LP}. Similar to GCN and GAT, we predict links via the Fermi-Dirac decoder. We also add another baseline named \textbf{PEGN (node-wise)}, which is a modification of our framework. We use the concatenation of the persistence images of the target nodes to replace the pair-wise topological feature in our framework.

\textbf{Experimental setup.} We adopt Cora, Citeseer, PubMed, Photo, Computers, and PPI as the benchmarks. Following~\citep{chami2019hyperbolic}, we use 5\% (resp.~10\%) of existing links in the input graph as the positive validation set (resp. positive test set). An equal number of non-existent links are sampled as the negative validation set and negative test set. The remaining 85\% existing links are used as the positive training set. In every epoch, we randomly choose the same number of remaining non-existent links as the negative training set. During training \textcolor{black}{and inference}, we remove positive validation and test edges from the graph. For fairness, we set the number of node embeddings of the hidden layer and the final layer to be the same (100 and 16) for all networks. The backbone GNN in our model is a classic 2-layer GCN with one hidden and one output layer. All the activation function used in the graph neural networks is RELU and all the activation function used in the Fermi-Dirac decoder is Leakyrelu with the negative slope set to 0.2. \textcolor{black}{The 2-layer MLP that combines the node feature and the persistence image is a fully-connected neural network with an input layer size of 41, a hidden layer size of 25, and a single output value.}

We use intersections of 1-hop neighborhoods as the vicinity graphs for PPI, Photo, and Computers. For Cora, Citeseer, and PubMed, we adopt intersections of their 2-hop neighborhood. Persistence images are all with dimension 25. Cross-Entropy Loss is chosen as the loss function and Adam is adopted as the optimizer with the learning rate set to 0.01 and weight-decay set to 0. Dropout is 0.8 for Cora and Citeseer, and 0.5 for the rest of the graphs. Notice that the dropout for Cora and Citeseer are different from our original work~\citep{yan2021link}, which is set to 0.5. The training epoch is 2000, and the early stopping on the validation set is 200 (patience epochs). The only exception is SEAL, due to its slow training speed and fast convergence, we only train 200 epochs. Following the settings in~\citep{zhang2018link,you2019position}, the RUC-AUC score is adopted as the evaluation metric. We run all the methods on each dataset 10 times to get the final results.

\textbf{Results.} Table~\ref{tab:LP} and Table~\ref{tab:PPI} report the results of all the methods. Since existing works either run too slow or do not converge on PPI, we do not list them in Table~\ref{tab:PPI}. Besides, we only list the performance of PEGN (FF) on some of the datasets due to its low speed. 

Compared with the baseline methods, our proposed framework performs comparably on Citeseer, and achieves the best performance on the other datasets. The results also indicate that the high-order topological feature is more effective on large and dense graphs, which has the same tendency as the experiments in node classification. 

Compared with models using node-wise persistent homology including PEGN (NC) and PEGN (node-wise), the proposed pair-wise framework generally performs better. This shows that the common structural information of the target nodes is beneficial for link prediction.

\subsection{End-to-end Learning of Localized Topological Features}
\label{subsec:exp_learn}

In short, we observe that the end-to-end learning of localized topological features does not contribute much in terms of node classification; while in link prediction benchmarks, it generally performs better. To explain the observation, GNNs may rely more on structural information to predict whether links exist or not. For node classification, the semantic information of graph elements may contribute more than the structural information. Detailed results are listed below.

\textbf{Node classification.} In Table~\ref{tab:NC}, \textbf{PEGN (PI)} and \textbf{PEGN (FF)} denote our proposed framework with ``combination of topological features" and ``learnable filter function" respectively. 
Comparing PEGN (FF) and PEGN (PI) with PEGN, we can find that PEGN generally achieves better results. This elucidates that Ollivier-Ricci curvature is indeed a suitable filter function for the framework. In addition, methods based on learnable persistent homology need additional parameters, which may lead to over-fitting.

\textbf{Link prediction.} We find that PEGN (PI) outperforms PEGN on larger datasets, which shows that the combination of persistence images extracts advantageous structural features on these graphs. This may originate from the rich and heterogeneous structures of these larger and denser datasets. PEGN (FF) achieves state-of-the-art results on PPI datasets. This shows that the learned topological feature may better suit these graphs. However, considering the much higher computational cost of the learnable filter function, it is under discussion whether it deserves to utilize the learnable filter function or not, especially on large graphs.

\section{Conclusion}

To enhance the power of GNNs on representation learning tasks, we propose a new topological feature based on the theory of persistent homology and integrate the feature with GNN models. With a specially designed vicinity graph and filter function, our approaches effectively extract the connectivity information between nodes and edges with theoretical guarantees, and achieve state-of-the-art results on various node classification and link prediction benchmarks. In addition, we discuss the structural information that persistent homology can capture, and the end-to-end learning of topological features. The paper can potentially serve as a guideline for further research in topological graph representation learning. 

\section*{Acknowledgments}
We thank Samantha Chen for her observation on the pair of graphs that 3-WL cannot differentiate. The work of Zuoyu Yan, Liangcai Gao, and Zhi Tang is supported by the projects of  Beijing Science and Technology Program (Z231100007423011) and National Natural Science Foundation of China (No. 62376012), which is also a research achievement of State Key Laboratory of Multimedia Information Processing and Key Laboratory of Science, Technology and Standard in Press Industry (Key Laboratory of Intelligent Press Media Technology).

\appendix

\section{Proof of Theorem~\ref{th:4wl}}

We restate the theorem as :

\begingroup
\def\thetheorem{\ref{th:4wl}}

\begin{theorem}
With the distance function~\citep{zhang2018link} as the filter function, Extended Persistence Diagrams can differentiate some pairs of graphs that 4-WL cannot differentiate.
\end{theorem}
\addtocounter{theorem}{-1}
\endgroup

\begin{proof}

\textcolor{black}{To begin the proof, we first formally define the 1-WL and $k$-WL algorithms~\citep{weisfeiler1968reduction}.}

\textcolor{black}{\textbf{1-WL.} The classic Weisfeiler-Leman algorithm (1-WL) is designed to compute a coloring for each node in the vertex set $V$ of a given graph $G$. The algorithm iteratively updates the color of each node by aggregating the color information from its neighbors. In the initial iteration (0-th iteration), the color assigned to a node $v \in V$ is denoted by $c_v^0$, representing its initial isomorphic type. For labeled graphs, the isomorphic type corresponds to the node's feature. For unlabeled graphs, all nodes are assigned the same initial color. In the $t$-th iteration, the color of node $v$ is updated as
$$c_v^t = \text{HASH}(c_v^{t-1}, \{\!\!\{c_u^{t-1} : u \in N(v)\}\!\!\}),$$
where HASH is a bijective hashing function that maps the input to a unique color. The process continues until the colors of all nodes remain unchanged between consecutive iterations. If two graphs yield different coloring histograms at the end (e.g., different numbers of nodes with the same color), 1-WL can distinguish them as non-isomorphic.}

\textcolor{black}{\textbf{k-WL.} For any $k \geq 2$, the $k$-dimensional Weisfeiler-Leman algorithm ($k$-WL) extends this approach by coloring $k$-tuples of nodes instead of individual nodes. In the initial iteration, the isomorphic type of a $k$-tuple $\vec{v}$ is determined by hashing: 1) the tuple of colors associated with the nodes in $\vec{v}$, and 2) the adjacency matrix of the subgraph induced by $\vec{v}$, ordered according to the node ordering within $\vec{v}$. In the $t$-th iteration, the coloring of $\vec{v}$ is updated as:
$$c_{\vec{v}}^t = \text{HASH}(c_{\vec{v}}^{t-1}, c_{\vec{v}, (1)}^{t}, \ldots, c_{\vec{v}, (k)}^{t}),$$
where 
$$c_{\vec{v}, (i)}^t = \{\!\!\{c_{\vec{u}}^{t-1} \mid \vec{u} \in N_i(\vec{v})\}\!\!\}, \quad i \in [k].$$ Here, $N_i(\vec{v}) = \{(v_1, \ldots, v_{i-1}, w, v_{i+1}, \ldots, v_k) \mid w \in V\}$ represents the $i$-th neighborhood of $\vec{v}$. Intuitively, $N_i(\vec{v})$ is obtained by replacing the $i$-th component of $\vec{v}$ with each node from $V$. Aside from the updating function, the other procedures of $k$-WL are analogous to those of 1-WL. In terms of graph distinguishing power, 2-WL is as powerful as 1-WL, and for $k \geq 2$, $(k+1)$-WL is strictly more powerful than $k$-WL.
}

\textbf{A pair of graphs that 4-WL cannot differentiate.} The constructed counter-example draws inspiration from the well-known Cai-F\"urer-Immerman (CFI) graph family~\citep{cai1992optimal}. We define a sequence of graphs $G^{(\ell)}, \ell = 0, 1, 2, 3, 4, 5$ as following,
	\begin{equation}
        \label{eq:cfi}
		\begin{split}
			V_{G^{(\ell)}} =& \Big\{u_{a, \vec{v}}\Big| a\in[5], \vec{v}\in\{0,1\}^4 \text{ and }\\
			&\quad\begin{array}{ll}
				\vec{v} \text{ contains an even number of } 1 \text{'s}, & \text{if }a=1,2,\ldots, 5-\ell, \\
				\vec{v} \text{ contains an odd number of } 1 \text{'s},  & \text{if }a=6-\ell,\ldots, 5.
			\end{array}\Big\}
		\end{split}
	\end{equation}
	Two nodes $u_{a,\vec{v}}$ and $u_{a',\vec{v}'}$ of $G^{(\ell)}$ are connected iff there exists $m\in [4]$ such that $a' \mod 5 = (a+m) \mod 5$ and $v_m = v'_{5-m}$. We define $G = G^{(0)} = (V_G, E_G)$ and $H = G^{(1)} = (V_H, E_H)$, which is a pair of non-isomorphic graphs that cannot be differentiated by 4-WL~\citep{yan2023efficiently}.

 \textbf{The chosen filter function.} In Section \ref{subsubsec:LP}, we have introduced the extraction of vicinity subgraphs for node pairs $v_1$ and $v_2$. Here, we formally define the filter function for nodes within these vicinity graphs. Specifically, each node $u$ in the vicinity subgraph of the node pair $(v_1, v_2)$ is assigned a label corresponding to its shortest path distances to $v_1$ and $v_2$:
 $$l(u) = (d(u,v_1),d(u,v_2))$$
 We then transform the function to a real number so that it can be adopted as the filter function for nodes:
 \begin{equation}
 \label{eq:filter_node}
 f(u) = 1 + min(d(u,v_1), d(u,v_2)) + (d/2)[(d/2) + (d\%2)-1],
 \end{equation}
 where $d = d(u,v_1) + d(u,v_2)$, $(d/2)$ and $(d\%2)$ are the integer quotient and remainder of $d$ divided by 2, respectively. The function computes different real numbers for nodes with different distances to $v_1$ and $v_2$.

Based on the definition of nodes, in the ascending filtration, the filter function for an edge $u_1u_2$ can be defined as:

\begin{equation}
\label{eq:filter_edge}
f(u_1u_2)=
\begin{cases}
max(f(u_1), f(u_2)) & \text{if } f(u_1) \neq f(u_2) \\
f(u_1) + 0.5 & \text{if } f(u_1) = f(u_2)
\end{cases}.
\end{equation}

\textcolor{black}{While in the descending filtration, the filter function for an edge $u_1u_2$ can be defined as:}

\begin{equation}
\label{eq:filter_edge_down}
f(u_1u_2)=
\begin{cases}
min(f(u_1), f(u_2)) & \text{if } f(u_1) \neq f(u_2) \\
f(u_1) - 0.5 & \text{if } f(u_1) = f(u_2)
\end{cases}.
\end{equation}

\textbf{EPD can differentiate the pair of graphs.} According to the definition of Equation~(\ref{eq:cfi}), we set the target nodes $v_1^G=u_{1, [0,0,0,0]}$ and $v_2^G=u_{2, [0,0,0,0]}$ in graph $G$. Similarly, we set $v_1^H=u_{1, [0,0,0,0]}$ and $v_2^H=u_{2, [0,0,0,0]}$ as the target nodes in graph $H$. Leveraging the filter function defined in Equation~(\ref{eq:filter_node}), we define $V_G^1$ as the set of nodes whose filter value is $2$: $$V_G^1 = \{f(u) = 2| u \in G\},$$ which is also the set of nodes whose shortest path distances to $v_1^G$ and $v_2^G$ are both $1$. An analogous set $V_H^1$ can be constructed for graph $H$ by replacing $G$ with $H$ in the definition.

Based on Equation~(\ref{eq:filter_edge}), edges whose filter values are $2.5$ indicate both incident nodes lie in $V_G^1$ (or $V_H^1$). Therefore, the number of the extended persistence points at coordinates $(2.5, 2.5)$ corresponds to the $1$-th Betti Number of the 
vicinity graphs $G_G^1 = (V_G^1, E_G^1)$ generated from $V_G^1$. Here the edge set $E_G^1 = (V_G^1)^2 \cap E_G$ consists of edges from $G$ with both endpoints in $V_G^1$. An analogous vicinity graph $G_H^1$ can be constructed for $V_H^1$.

In $G_G^1$, there exist two distinct 3-cycles (cycles with 3 nodes): $u_{5, [0,0,0,0]} u_{4, [0,0,0,0]} u_{3, [0,0,0,0]}$ and $u_{5, [0,0,1,1]}u_{3, [1,1,0,0]}u_{4, [1,0,0,1]}$, thus its $1$-th Betti Number is 2. In contrast, $G_H^1$ contains a single 6-cycle: $u_{5, [0,0,1,0]}u_{3, [1,1,0,0]}u_{4,[1,0,0,1]}u_{5, [0,0,0,1]}u_{3, [0,0,0,0]}u_{4,[0,0,0,0]}$, thus its $1$-th Betti Number is 1. Consequently, based on the defined filter function, the EPD can differentiate the pair of graphs since all nodes are treated as the same in these two graphs.

\end{proof}

\section{Proof of Theorem~\ref{th:ph_4wl}}

We restate the theorem as :

\begingroup
\def\thetheorem{\ref{th:ph_4wl}}
\begin{theorem}
With the shortest path distance as the filter function, the 1D Extended Persistence Diagram is less powerful than 4-WL in terms of differentiating non-isomorphic graphs.
\end{theorem}
\addtocounter{theorem}{-1}
\endgroup
\begin{figure}
\begin{minipage}{\linewidth}
\begin{algorithm}[H]
	\begin{algorithmic}[1]
	\STATE {\bfseries Input:} filter function $f$, input graph $G = (V,E)$
	\STATE $V, E = \text{sorted} (V, E, f)$, $PD_1 = \{\}$
	\FOR{$i \in V$}
	\STATE $C_i = \{C_{i_j} | (i, j) \in E, f(j) > f(i)\}$, $E_i = E$
    \FOR{ $C_{i_j} \in C_i$}
	\STATE $f(C_{i_j}) = f(i)$, $E_i = E_i - \{(i,j)\} + \{(C_{i_j}, j)\}$
	\ENDFOR
	\STATE $PD_1^i = \text{Union-Find-step}(V + C_i - \{i\}, E_i, f, C_i)$
	\STATE $PD_1 += PD_1^i$
	\ENDFOR
	\STATE {\bfseries Output:} $PD_0$, $PD_1$
	\end{algorithmic}
    \caption{Computation of 1D EPD corresponding to cycles}
	\label{alg:ext_PD}
    \end{algorithm}

\end{minipage}
\begin{minipage}{\linewidth}
		\begin{algorithm}[H]
		\begin{algorithmic}[1]
		\STATE {\bfseries Input:} $V$, $E$, $f$, $C_i$ 
		\STATE $PD_1^i = \{\}$
		\FOR{$v \in V$}
		\STATE $v.value = f(v)$, $v.root = v$ 
		\ENDFOR
		\STATE $Q = \text{Sort}(V)$, $Q = Q - \{v|f(v) < f(i)\}$, $G = \{Q, E_Q\}$, where $E_Q = E \cup Q^2$.
		\WHILE{Q is not empty}
		\STATE $u = Q.\text{pop-min}()$
		\FOR{$v \in G.\text{neighbors}(u)$}
		\STATE $pu, pv = \text{Find-Root}(u), \text{Find-Root}(v)$
		\IF{$pu \neq pv$}
		\STATE $s = argmin(pu.value, pv.value)$
		\STATE $l = argmax(pu.value, pv.value)$
		\STATE $l.root = s$ 
		\IF{$pu \in C_i$ and $pv \in C_i$} 
		\STATE $PD_1^i + \{(u.value, l.value)\}$
		\ENDIF 
		\ENDIF
		\ENDFOR
		\ENDWHILE
		\STATE {\bfseries Function:} $\text{Find-Root} (u)$
		\STATE $pu = u$
		\WHILE {$pu \neq pu.root$}
		\STATE $pu.root = (pu.root).root$, $pu = pu.root$
		\ENDWHILE
		\STATE {\bfseries Return:} $pu$
		\end{algorithmic}
		\caption{Union-Find-step}
		\label{alg:UFs}
		\end{algorithm}
		\end{minipage}
\end{figure}

\begin{proof}
The filter function can be built using Equation~(\ref{eq:filter}).

\textbf{Computation of EPDs with regard to Algorithm~\ref{alg:ext_PD}.} We present the computation of the extended persistence diagrams (EPDs) using the algorithm described in~\citep{yan2022neural}. The algorithm is presented in Algorithm~\ref{alg:ext_PD}. We aim to demonstrate that each extended persistence point in the EPD can be captured by a specific model called subgraph WL which is rooted on 2-tuples, such as~\citep{huang2023boosting}. Firstly, we build the shortest path tree and the filter function on the root node $w$. For each node $i \in V$, we construct a subgraph with a node set containing only those nodes whose filter values are greater than $f(i)$, and an edge set consisting only of edges whose both filter values are greater than $f(i)$. The subgraph corresponds to the subgraph made by $E_i$ and $V+V_i-{i}$ in Algorithm~\ref{alg:ext_PD}.

\textbf{Each step in Algorithm~\ref{alg:UFs}.} In the extracted subgraph, we assign a unique label to each edge that contains node $i$, respectively. This set of labels is denoted as $L = \{l_{i_1}, \dots, l_{i_p}\}$, where $\{i_1, \dots, i_p\}$ are the neighbors of node $i$ in the subgraph. Subsequently, we employ an MPNN-like model on the subgraph to derive the extended persistence points. For initialization (the $0$-th iteration), we label only the edges containing $i$ using their corresponding unique labels: $h^0_{(i, i_t)} = l_{i_t}$, where $t = 1, \dots, p$. In the $k$-th iteration, we have:
\begin{equation}
h^k_{u} = \text{AGGR}\{h^{k-1}_{(v_1, u)} |(v_1, u) \in E^{k, 1}_u\}
\end{equation}
and 
\begin{equation}
h^k_{(u, v_2)} = h^k_{u}, \text{ for each } (u, v_2) \in E^{k, 2}_u
\end{equation}
where $u$ represents an unlabeled node at iteration $k-1$, which is connected to edges labeled in iteration $k-1$. We define $E_u^{k,1}$ as the set of labeled edges connected to node $u$ and $E_u^{k,2}$ as the set of unlabeled edges connected to node $u$. The set-level aggregation function $AGGR$ outputs the first label within a set. If there are $q$ distinct labels in ${h^{k-1}_{(v_1, u)} |(v_1, u) \in E^{k, 1}_u}$, the model produces $q-1$ extended persistence points $(k, f(i))$. By combining all the extended persistence points in the subgraph, we can obtain the EPD of the subgraph, which is also the output of Algorithm~\ref{alg:UFs}. 

\textbf{Subgraph-WL can approximate the computation.} We then show that such computations can be approximated by a subgraph WL rooted on 2-tuples. Using the definition of~\citep{yan2023efficiently}, for a given 2-tuple $(w, i)$ and the corresponding subgraph $G_{(w, i)} = (V_{(w, i)}, E_{(w, i)})$, the subgraph WL is updated by:
\begin{equation}
\begin{split}
h^t_{u, (w, i)} = \text{HASH} \{h^{t-1}_{u, (w, i)}, h^{t-1}_{w, (w, i)}, h^{t-1}_{i, (w, i)},  \\
\{h^{t-1}_{v, (w, i)} | (u, v) \in E_{(w, i)}\}\}
\end{split}
\end{equation}

Notice that the set of all filter values forms a countable set, therefore there exists a bijective hashing function ``HASH" that can approximate the computation of the 1D EPDs. Specifically, it can be represented as:
\begin{equation}
\begin{split}
\text{HASH} \{h^{t-1}_{u, (w, i)}, h^{t-1}_{w, (w, i)}, h^{t-1}_{i, (w, i)}, \{h^{t-1}_{v, (w,i)} | (u, v) \in E_{(w, i)}\}\}= \\
\begin{cases}
h^{t-1}_{u, (w, i)} & \text{if } h^{t-1}_{u, (w, i)} \text{ has label} \\
(\text{the first } h^{t-1}_{v, (w, i)}, \text{Hist}(\{h^{t-1}_{v, (w,i)}\})) & \text{otherwise}
\end{cases}.
\end{split}
\end{equation}
where ``Hist" is the label histogram extracted by the ``HASH" function. In addition, the 0D ordinary persistence diagram can be computed by a Union-Find algorithm, which can be approximated by a subgraph WL similarly. Considering that the subgraph WL is less powerful than 4-WL~\citep{huang2023boosting, yan2023efficiently, frasca2022understanding}, the EPDs are less powerful than 4-WLs in terms of distinguishing non-isomorphic graphs\footnote{\textcolor{black}{The result applies to both using the original lower-star and upper-star filtration~\citep{cohen2009extending} or the ascending and descending filtration adopted in the proof of Theorem~\ref{th:ph_3wl}.}}.

\end{proof}

\section{Proof of Theorem~\ref{th:ph_regular}.}
We restate the theorem as:
\begingroup
\def\thetheorem{\ref{th:ph_regular}}
\begin{theorem}
Consider all pairs of $n$-sized $r$-regular graphs, let $ 3 \leq r < (2log2n)^{1/2}$ and $\epsilon$ be a fixed constant. With at most $K=\lfloor(\frac{1}{2}+\epsilon)\frac{\log{2n}}{\log{(r-1)}} \rfloor$, there exists an Extended Persistence Diagram with the shortest path distance as the filter function that distinguishes $1-o(n^{-1/2})$ such pairs of graphs.
\end{theorem}
\addtocounter{theorem}{-1}
\endgroup

\begin{proof}

We have demonstrated that by utilizing the filter function defined in Equation~(\ref{eq:filter}), we can extract three important structural features from a graph: (1) the number of nodes within layer $k$; (2) the number of edges connecting layer $k$ to layer $k+1$; and (3) the number of edges within layer $k$. Given an input graph $G = (V, E)$, we denote the root node of the filter function as $v$, and we represent the concatenation of these three numbers as $EPD^k_{v, G}$.

We then introduce the following lemma.
\begin{lemma}
\label{lem:node_configuration}
For two graphs $G^{(1)}=(V^{(1)},E^{(1)})$ and $G^{(2)}=(V^{(2)},E^{(2)})$ that are randomly and independently sampled from $n$-sized $r$-regular graphs with $ 3 \leq r < (2log2n)^{1/2}$. We pick two nodes $v_1$ and $v_2$ from two graphs respectively.  Let $K=\lfloor(\frac{1}{2}+\epsilon)\frac{\log{2n}}{\log{(r-1)}} \rfloor$ where $\epsilon$ is a fixed constant, $EPD^{K}_{v_1,G^{(1)}}=EPD^{K}_{v_2,G^{(2)}}$ with the probability at most $o(n^{-3/2})$.
\end{lemma}

The proof of this Lemma can be derived from Theorem 6 in~\citep{BOLLOBAS198233} and Lemma 1 in~\citep{feng2022how} with minor modifications, and we present it below.

Here, the configuration model~\citep{bollobas_probabilistic_1980} is introduced as a means of generating $n$-sized $r$-regular graphs. Suppose we have $n$ disjoint sets of items $W_i$, each of which contains $r$ items that correspond to nodes in the model. A configuration is a partition of all $nr$ items into $\frac{nr}{2}$ pairs, and $\Omega$ represents the set of configurations. Among all configurations in $\Omega$, given $r< (2logn)^{1/2}$, there exist approximately $\exp(-\frac{r^2-1}{4})$ or $\Omega ( n^{-1/2})$ portion of them are simple $r$-regular graphs. For these configurations, there exists an edge between node $i$ and $j$ in the corresponding $r$-regular graph if there exists a pair of items with one item from set $W_i$ and another item from set $W_j$.

Consider a pair of nodes $i$ and $j$ randomly selected from the configuration model of an $n$-sized $r$-regular graph, and let $l_0=\lfloor(\frac{1}{2}+\epsilon)\frac{\log{n}}{\log{(r-1)}} \rfloor$. To generate a graph, we select all edges that directly connect to the two nodes in the first step and obtain all nodes that are at a distance of $1$ to either of them. We then select all edges that connect to either $i$ or $j$ at a distance of $1$ in the second step and continue this process iteratively for $n-1$ steps until we have the union of components of the two nodes in a random configuration.

In the configuration, dispensable edges can occur under two potential conditions: 1) when the two ends of the edge are either $i$ or $j$; 2) when both ends of the edge are connected to at least one other edge. Since the first $k-1$ edges selected can connect to at most $k+1$ nodes, the probability that the $k$-th edge selected is dispensable can be upper-bounded as follows:

\begin{equation*}
\frac{(k+1)(r-1)}{(n-k-1)r} \approx \frac{k}{n-k}.
\end{equation*}

Thus, the probability of selecting more than two dispensable edges among the first $k_o=\lfloor n^{1/6} \rfloor$ edges is upper bounded as follows:

\begin{equation}
\label{eq:dispensable1}
\binom{k_o}{3} {\left(\frac{k_o}{n-k_o}\right)}^3=o(n^{-2}).
\end{equation}

The probability of selecting more than $l_1=\lfloor n^{1/8}\rfloor$ dispensable edges among the first $k_1=\lfloor n^{6/13}\rfloor$ edges is upper bounded as follows:

\begin{equation}
\label{eq:dispensable2}
\binom{k_1}{l_1+1} \left( \frac{k_1}{n-k_1}\right)^{l_1+1}=o(n^{-2}).
\end{equation}

The probability of selecting more than $l_2=\lfloor n^{5/13 }\rfloor$ dispensable edges among the first $k_2=\lfloor n^{2/3}\rfloor$ edges is upper bounded as follows:

\begin{equation}
\label{eq:dispensable3}
\binom{k_2}{l_2+1}\left( \frac{k_2}{n-k_2}\right)^{l_2+1}=o(n^{-2}).
\end{equation}

Let $A$ be the event that at most 2 of the first $k_o$ edges are dispensable, at most $l_1$ of the first $k_1$ edges are dispensable, and at most $l_2$ of the first $k_2$ edges are dispensable. Using Equations~(\ref{eq:dispensable1})-(\ref{eq:dispensable3}), it follows that the probability of $A$ is $1-o(n^{-2})$.

Event $A$ ensures that in the early steps of the graph generation process, the edges connect to almost as many nodes as possible and the number of nodes at each distance ${1, 2, ..., k}$ from nodes $i$ and $j$ is $r, r(r-1), ..., r(r-1)^{k-1}$ for their respective sets $W_i$ and $W_j$. This implies that $EPD^{k}_{i,G}=EPD^{k}_{j,G}$ is very likely to hold with probability close to 1. Later nodes in the configuration have different properties.

We now consider an alternative approach for generating a graph using the configuration model. Similar to the previous approach, we begin by selecting all the edges that connect to at least one node with a distance less than $k$ from $i$ and $j$ after the $k$-th step. At the $(k+1)$-th step, we first select all the edges with one endpoint at a distance of $k$ from node $i$. Next, we select all the edges with both endpoints at a distance of $k$ from node $j$. It is evident that after these two procedures, the only edges remaining to complete the $(k+1)$-th step are those that have one endpoint at a distance of $k$ from node $j$ and the other endpoint at a distance of $k+1$ from node $j$. Let $t_k$ denote the number of edges in the nodes at distance $k$ from node $j$ that have not been generated yet, and $s_k$ denote the number of nodes that have not generated any edge so far. The final step to complete the $(k+1)$-th step is to connect the $t_k$ edges to the $s_k$ nodes. This procedure is repeated for every $k$ to generate the final graph.

Assuming event $A$ occurs, for $k \leq l_0$, the following inequality holds:

\begin{equation}
\label{eq:tksk}
t_k\geq (r-1)^{k-3}  \quad \textrm{and} \quad  s_k\geq n/2,
\end{equation}

After the first two procedures of $k+1$-th step, we have all the nodes at the distance of $k+1$ from node $i$, which means we already computed $EPD^{k+1}_{i,G}$. Denote the number of $k$-hop neighbors from of node $i$ in $G$ as $N_{i,G}^k$. If $N^{k+1}_{i,G}$ is the same as $N^{k+1}_{j,G}$, then the connection of $t_k$ edges must belong to those $N_{i,G}^k$ nodes from totally $s_k$ nodes. The probability of $N^{k+1}_{i,G}=N^{k+1}_{j,G}$ condition on Equation~(\ref{eq:tksk}) is at most the maximum of the probability that 
$t_k$ edges connect to $l$ nodes from $s_k$ nodes with degree $r$. This probability is bounded by:

\begin{equation}
\label{eq:subset}
\underset{l}{max}\ P(N^{k+1}_{j,G}=l)\leq c_o \frac{s_k^{1/2}}{t_k},
\end{equation}

where $r\geq 3$ and $t_k\leq cs_k^{5/8}$ for some constant $c$ and $c_o$ is also a constant. Such proof can be found in Lemma 7 of~\citep{bollobas_probabilistic_1980}. Given Equation~(\ref{eq:subset}), the probability that $EPD^{l}_{i,G}=EPD^{l}_{j,G}$ for $l \leq l_0$ is at most

\begin{equation*}
1-P(A)+\prod^{l_0}_{l=h} c_0\frac{n^{1/2}}{(r-1)^{l-3}},
\end{equation*}

where $h=\lfloor \frac{1}{2} \frac{log n}{log (r-1)}\rfloor +3$. Since $(r-1)^{l_0} \geq n^{(1+\epsilon)/2}$, the sum above is $o(n^{-2})$. Moreover, as there are at least $\Omega(n^{-1/2})$ of all the graphs generated by the configuration model that are simple $r$-regular graphs, there are at most $o(n^{-2}/n^{-1/2})=o(n^{-3/2})$ probabilities that $EPD_{i,G}^{l_0} = EPD_{j,G}^{l_0}$.

We can combine any pair of $n$-sized $r$-regular graphs $G^{(1)}$ and $G^{(2)}$ to generate a single regular graph $G_c$ with $2n$ nodes. The proof presented above for $G$ also holds for $G_c$. This implies that if we randomly select a node $v_1$ from the first component and a node $v_2$ from the second component of $G_c$, then given $3\leq r<(2\log{2n})^{1/2}$ and $K=\lfloor(\frac{1}{2}+\epsilon)\frac{\log{2n}}{\log{(r-1)}} \rfloor$, we have $EPD^{K}_{v_1,G^1} = EPD^{K}_{v_1,G_c}=EPD^{K}_{v_2,G_c} = EPD^{K}_{v_2,G^2}$ with probability of $o(n^{-3/2})$.
\end{proof}

\textcolor{black}{\section{Discussion on the choice of filter function}
\label{subsec:filter_discuss}
In Theorem~\ref{th:ph_3wl}, we facilitate the proof by introducing a relaxation function, as shown in Equation~(\ref{eq:filter}). However, the relaxed function lacks elegance, so we adopt the original shortest-path-distance-based function in our paper. To verify this choice, we conducted an additional experiment where the filter function used in the proof of Theorem~\ref{th:ph_3wl} was applied. The results are presented in Table~\ref{tab:NC1}, with the model PEGN using the new filter function denoted as "PEGN (new)." Our findings indicate that the model's performance remains nearly identical to that with the original filter function. This stability can be attributed to the robustness of EPDs~\citep{cohen2007stability, cohen2009extending}. Given that the changes to the filter functions are minimal (we can, in fact, assign a distinct value significantly smaller than 0.5), the corresponding EPDs and persistence images are only slightly affected, leading to a negligible impact on the model's performance.
\begin{table*}
	\centering
	\caption{Discussion on the choice of filter function on node classification benchmarks.}
	\label{tab:NC1}
	\scalebox{0.9}{
		\begin{tabular}{lccccccc}
			\hline\noalign{\smallskip}
			Method & Cora & Citeseer &  PubMed & CS & Physics & Computers & Photo \\
			\noalign{\smallskip}\hline\noalign{\smallskip}
			PEGN  & 82.7$\pm$0.4 & 71.9$\pm$0.5 & 79.4$\pm$0.7 & 93.3$\pm$0.3& 94.3$\pm$0.1 & 86.6$\pm$0.6 & 92.7$\pm$0.4\\
                PEGN (new)  & 82.6$\pm$0.5 & 71.9$\pm$0.5 & 79.5$\pm$0.9 & 93.3$\pm$0.3 & 94.4$\pm$0.1 & 86.6$\pm$0.6 & 92.7$\pm$0.4\\
        \noalign{\smallskip}
			\hline
			\noalign{\smallskip}
	\end{tabular}}
\end{table*}
}

\bibliography{bare_jrnl_compsoc}
\end{document}